\documentclass[11pt]{article}

\usepackage[preprint]{acl}

\usepackage{times}
\usepackage{latexsym}

\usepackage[T1]{fontenc}

\usepackage[utf8]{inputenc}

\usepackage{microtype}

\usepackage{inconsolata}

\usepackage{graphicx}
\usepackage{bm}
\usepackage{enumitem}
\usepackage{amsthm}
\usepackage{setspace}
\usepackage{amsmath}
\usepackage{amssymb}
\usepackage{booktabs}
\usepackage{multirow}
\usepackage{booktabs}
\usepackage[table]{xcolor}
\usepackage{xspace}
\usepackage{tcolorbox}
\usepackage{pifont}
\usepackage{caption}
\usepackage{longtable}
\usepackage{colortbl}
\usepackage{array}
\usepackage{booktabs}
\usepackage{tabularx}
\usepackage{cellspace}
\usepackage{setspace}
\usepackage{listings}
\usepackage{inconsolata}
\tcbuselibrary{skins}
\tcbuselibrary{listings}
\tcbuselibrary{breakable}
\newcommand{\name}{\textsc{SocaSim}\xspace}

\definecolor{myblue}{RGB}{8, 69, 148}
\definecolor{mylightblue}{RGB}{222, 235, 247}
\definecolor{mycolor}{RGB}{247, 251, 255}

\tcbset{
  mybox/.style={
    colback=gray!10,      
    colframe=gray!20,     
    coltitle=black,      
    fonttitle=\bfseries\large\setstretch{1.2},
    rounded corners=all,  
    boxrule=1pt,         
    left=10pt,           
    right=10pt,          
    top=5pt,              
    bottom=5pt
  }
}

\tcbset{
  myboxblue/.style={
    colback=mylightblue, 
    colframe=myblue,     
    coltitle=white,          
    colbacktitle=myblue,      
    fonttitle=\bfseries\large\setstretch{1.2},
    rounded corners=all,
    boxrule=1pt,
    left=10pt,
    right=10pt,
    top=5pt,
    bottom=5pt
  }
}

\title{From Blueprint to Reality: Modeling and Applying Putnam’s Social Capital Theory with LLM-based Multi-agent Simulations}

\author{
  \textbf{Shiyi Ling\textsuperscript{1}},
  \textbf{Zhi Zheng\textsuperscript{1}},
  \textbf{Hui Zheng\textsuperscript{2}},
  \textbf{Wenjun Xue\textsuperscript{3}},
  \textbf{Feng Ye\textsuperscript{4}},
  \textbf{Tong Xu\textsuperscript{1}}
  \\
\textsuperscript{1}State Key Laboratory of Cognitive Intelligence, University of Science and Technology of China
\\
\textsuperscript{2}Anhui University
\\
\textsuperscript{3}North Automatic Control Technology Institute
\\
\textsuperscript{4}University of Science and Technology of China
\\
\small{
\href{mailto:shiyi.ling@mail.ustc.edu.cn,lawaken215@mail.ustc.edu.cn, yefengustc@mail.ustc.edu.cn}{\{shiyi.ling, awaken215, yefengustc\}@mail.ustc.edu.cn}, \href{mailto:zhengzhi97@ustc.edu.cn,tongxu@ustc.edu.cn}{\{zhengzhi97, tongxu\}@ustc.edu.cn},
    \href{mailto:huizheng@ahu.edu.cn}{huizheng@ahu.edu.cn},
  }
}

\begin{document}
\maketitle
\begin{abstract}
Putnam's Social Capital Theory is a foundational framework for collective action and community prosperity. However, traditional empirical methods face practical limits on control and replication. Meanwhile, LLM-based social simulations are typically behavior‑driven and lack theory‑aligned environments for modeling Putnam's core propositions. To address these gaps, we introduce {\bf \name{}}, an LLM-based multi-agent simulation framework to study Putnam's Social Capital Theory from theoretical blueprint to simulated reality. Specifically, we build an environment integrating social network evolution, trust dynamics, and norm propagation, where agents engage in repeated collective-action experiments, and then apply the three dimensions to analyze adaptation challenges in smart elderly care. Our simulations reproduce Putnam's macro-level patterns and exhibit strong human-agent alignment at the group level. Unlike traditional methods, {\bf \name{}} traces micro-level causal pathways of social network, trust, and norms via round-by-round simulations and counterfactual interventions, enabling process-level interpretability. Taken together, these capabilities establish a research paradigm that leverages LLM agents to bridge social science and computer science.
\end{abstract}

\section{Introduction}

When an accident occurs, large anonymous crowds often freeze, each waiting for someone else to move first. In close‑knit groups, help starts immediately because people are connected, trust one another, and expect to assist. This contrast shows how social network, trust, and norms shape a group's ability to respond to common problems. Social Capital Theory, developed by Robert D. Putnam \citep{putnam1993,putnam2000}, captures these dynamics and explains why some communities overcome collective‑action dilemmas while others do not.

\begin{figure}[ht]
\vspace{-0.8\baselineskip}
  \vskip 0.2in
  \begin{center}
    \centerline{\includegraphics[width=\columnwidth]{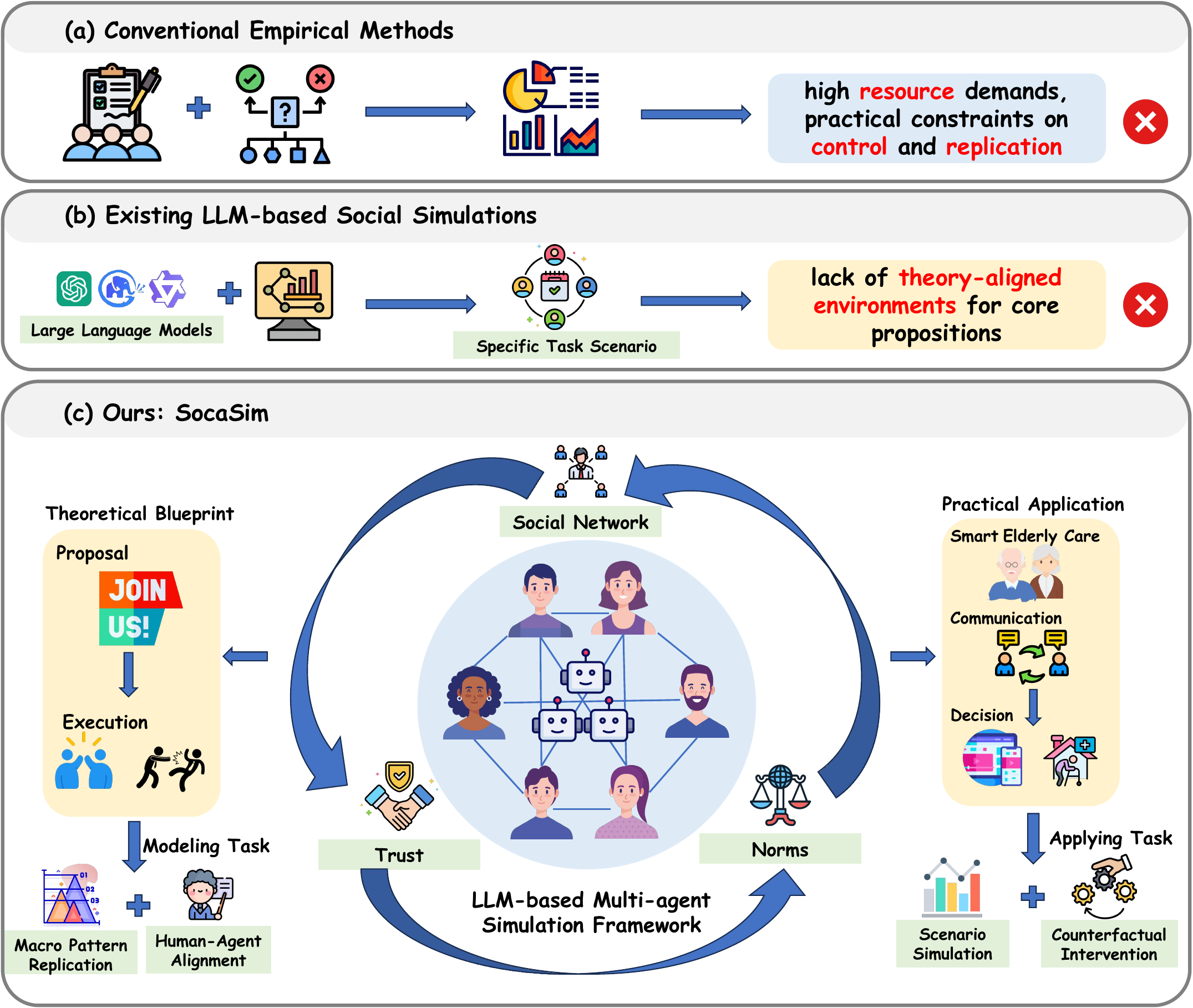}}
    \caption{
      Comparison of existing research paradigms versus {\bf \name{}} for Putnam’s Social Capital Theory.
    }
    \label{Figure 1}
  \end{center}
  \vspace{-2\baselineskip}
\end{figure}

In social science, research on Putnam’s Social Capital Theory largely relies on conventional empirical methods that yield valuable insights yet have notable limits (Figure~\ref{Figure 1}(a)). Quantitative methods such as large-scale surveys \citep{durante2024} and Structural Equation Modeling (SEM) \citep{hoa2021,sumi2025} require substantial time and resources, and face practical constraints on control and replication. In parallel, simulation-based approaches such as Agent-Based Modeling (ABM) \citep{shenk2019} allow controlled tests of assumptions, yet traditional rule-based agents rely on predefined and simple decision rules and struggle to capture the complexity of human reasoning, emotion, and context-dependent behavior.

Recently, advances in large language models (LLMs) have enabled the simulation of human‑like intelligence \citep{friha2024,wang2025}, prompting growing interest in LLM‑based agents for social simulation \citep{piao2025,yang2025}. However, existing approaches are largely \textit{behavior‑driven} rather than \textit{theory‑driven} (Figure~\ref{Figure 1}(b)): they typically test whether agents exhibit plausible or human‑like behavior on specific tasks \citep{Zhao2024,Jia2025,Xie2024}, but rarely construct environments aligned with core theoretical propositions. Furthermore, current frameworks seldom provide reproducible, controllable settings for such theory‑driven analysis, for example, by modeling links between network connectivity and collective outcomes or stepwise trust growth. Consequently, they still make it difficult to trace how trust accumulates, how norms are internalized, or how individual–collective tensions arise, leading to limited procedural interpretability.

To bridge this gap, we introduce {\mbox {\bf \name{}}}, a dynamic and theory‑grounded LLM-based multi-agent simulation framework designed to model and apply Putnam’s Social Capital Theory (Figure~\ref{Figure 1}(c)). Specifically, we construct an agent society that integrates mechanisms of dynamic social network evolution, trust adaptation, and norm diffusion, in which agents engage in repeated collective-action simulations. Each agent is endowed with basic demographic attributes and social capital endowments, and is capable of reasoning and decision-making by incorporating social relationships, interaction history, and situational context. Our agent society runs in rounds with two phases: \textit{Proposal}, where agents make proposals; and \textit{Execution}, where they simultaneously decide whether to act based on trust, norms, and network ties.

Based on this framework, we conduct \textbf{two progressive tasks}. \textit{In the modeling task}, we run \textbf{multi-round collective-action experiments and human-alignment validation}. The simulation reproduces the macro-level patterns predicted by Putnam’s theory, and a comparison with real older adults shows strong human-agent alignment in group-level decisions (Pearson r = 0.974). Subsequently, \textit{in the applying task}, we move the theory from blueprint to reality by \textbf{applying its three core dimensions to adaptation challenges in smart elderly care}, where technology adoption hinges on social networks, trust, and norms. In counterfactual simulations that raise low-SES agents’ initial trust, the adoption rate increases by 15.4\% and decision contradictions decrease by 25.5\%, suggesting that trust can serve as a key causal lever. Unlike traditional methods that are often limited to correlational conclusions, {\bf \name{}} uses LLM agents to reveal micro-level causal chains of how trust accumulates, norms are internalized, and decision contradictions emerge, offering process‑level interpretability and practical guidance for policy design.

Our contributions can be summarized as follows:

\noindent $\bullet$ We propose {\bf \name{}}, a new interdisciplinary research paradigm using LLM‑based agents to model Putnam's Social Capital Theory.

\noindent $\bullet$ We develop a multi-agent framework that models the dynamics of social capital, including social network change, trust dynamics, and norms diffusion.

\noindent $\bullet$ We conduct several experiments with macro‑level replication, human alignment, and counterfactual intervention, validating {\bf \name{}}'s explanatory power for theory and real-world issues.

\begin{figure*}[ht]
  \vskip 0.2in
  \begin{center}
    \centerline{\includegraphics[width=\textwidth]{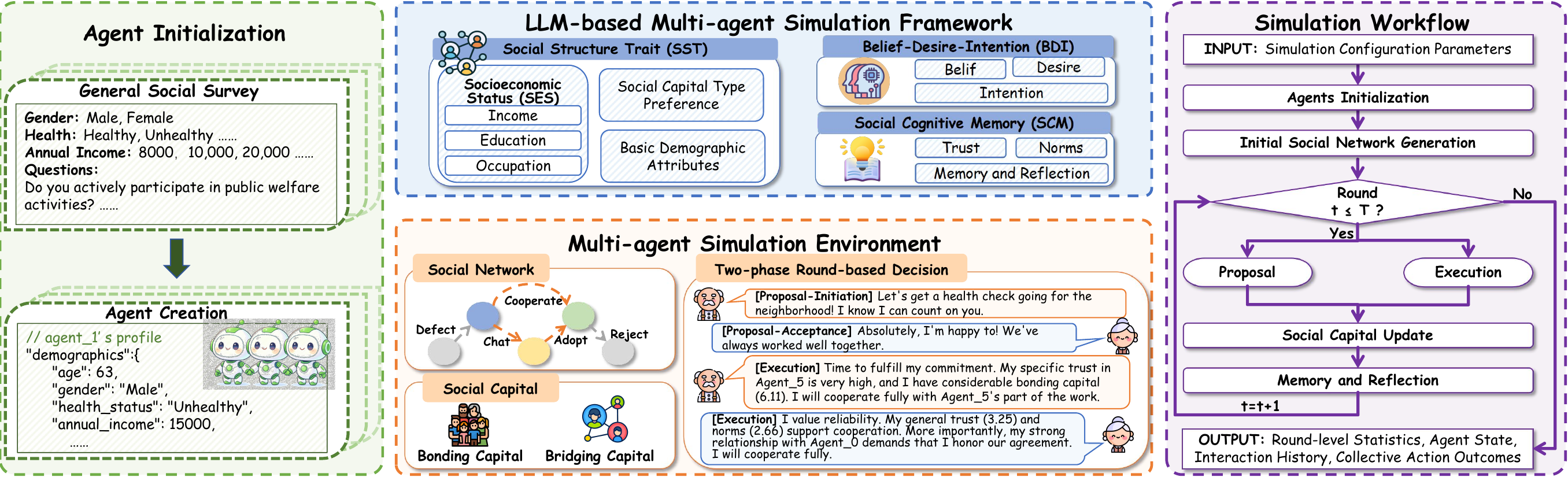}}
    \caption{
      Overview of {\bf \name{}} multi-agent architecture and round‑by‑round simulation workflow.
    }
    \label{Figure 3}
  \end{center}
  \vspace{-2\baselineskip}
\end{figure*}

\section{Preliminary}

To clarify the theoretical basis, we briefly introduce Putnam’s Social Capital Theory (see Appendix~\ref{app:preliminary} for more examples). This theory is commonly used to explain how groups deal with \textit{collective-action dilemmas}, namely situations where a shared problem is hard to resolve because coordination is costly, trust is insufficient, or long-term participation is hard to sustain. Putnam argues that \textit{social capital} helps alleviate such dilemmas and improves a group’s ability to respond to common challenges.

Putnam distinguishes between \textit{two primary forms} of social capital. \textbf{Bonding social capital} develops within close‑knit groups and strengthens internal support and solidarity. \textbf{Bridging social capital} connects different or loosely linked groups and facilitates information exchange and access to external resources. Conceptually, social capital can be understood through \textit{three interrelated dimensions}: \textbf{social network}, which provides channels for social interaction and coordination; \textbf{trust}, which reduces uncertainty and lowers the perceived cost of collective response; and \textbf{norms}, which encourage reciprocity, mutual support, and sustained participation within the group. \textbf{These dimensions correspond directly to the core components of our simulation}: social network shapes interaction opportunities, trust guides expectations about others’ actions, and norms regulate behavior over time.

Based on these two primary forms and three core dimensions, we leverage LLM agents to model and apply Putnam's Social Capital Theory.

\section{Methods}
\subsection{Overview of {\bf \name{}}}

Since Putnam's Social Capital Theory involves complex social constructs (social network, trust, norms), prior general multi-agent frameworks cannot capture their dynamic coupling. We therefore propose {\bf \name{}}, an LLM‑based multi‑agent framework to model, dynamically couple, and validate the theory's core elements. Specifically, we construct a social simulation system consisting of $N$ agents, denoted as $A = \{a_1, \dots, a_N\}$, each equipped with demographic and socioeconomic attributes and initial social‑capital endowments, situated in a controlled environment $E$. The simulation runs for $T$ discrete rounds. In each round $t$ ($t = 1, \dots, T$), agents initiate or respond to social behaviors based on their network connections, trust dispositions, and perceived norms. These interactions update three interrelated core dimensions:  

\noindent $\bullet$ \textbf{Social network $\bm{G_{t}}$}: Connections dynamically form, strengthen, or weaken over time based on interaction frequency and quality.

\noindent $\bullet$ \textbf{Trust $\bm{T_{i}(t)}$}: Each agent $a_{i}$ evolves its level of \textit{specific trust} toward specific others and \textit{general trust} toward the broader group through interactions.

\noindent $\bullet$ \textbf{Norms $\bm{N_{i}(t)}$}: Each agent’s reciprocity norms comprise \textit{specific reciprocity} for direct resource exchange and \textit{general reciprocity} for mutual cooperation without expecting immediate return.

As shown in Figure~\ref{Figure 3}, these abstract dimensions are concretely implemented by the agent architecture (\textit{i.e.}, SST, BDI, and SCM) and the LLM‑driven updating mechanisms detailed in \S\ref{sec:framework}.

\subsection{LLM-based Multi-agent Framework} \label{sec:framework}
\subsubsection{Social Structure Trait (SST)}

Putnam's Social Capital Theory holds that socioeconomic position determines access to networks and resources, yet most multi-agent frameworks lack ways to model this. To address the limitation, we design a module that generates agents' structural social features through three key components:

\noindent \textbf{Basic Demographic Attributes.} Each agent is assigned demographic traits, including age, gender, education, income, health status, and occupation. These attributes are sampled from the 2023 Chinese General Social Survey (CGSS) data\footnote{\url{https://www.cnsda.org}} elderly subsample, ensuring that the simulated population reflects real-world socioeconomic stratification.

\noindent \textbf{Socioeconomic Status (SES).} A weighted composite of income, education, and occupation, classifying agents into low, mid, and high SES categories.

\noindent \textbf{Social Capital Type Preference.} Low‑SES agents tend to rely on bonding capital, high‑SES agents are more adept at leveraging bridging capital, while mid‑SES agents exhibit a balanced preference.

All these attributes are encoded into the agent's natural language prompt, giving each agent a rich identity and ensuring behavior arises from sociologically plausible profiles rather than static rules.

\subsubsection{Belief-Desire-Intention (BDI)}

To capture the real-time decision-making of agents based on structural features from the SST module, we adopt the BDI framework \citep{bratman1999}. Specifically, we formalize the agent's context-sensitive cognitive decision process, with 
$$
\text{Act} \gets \mathrm{BDI}(f, B, D, I, S, M),
$$
where $\text{Act}$ denotes the final action, $f$ is implemented with an LLM, and $B, D, I, S, M$ denote the \textit{Belief}, \textit{Desire}, \textit{Intention}, SST structures, and SCM memory, respectively. \textit{Belief} combines structural cues such as SES and education with situational cues from the current interaction round. \textit{Desire} represents motivations such as technological mastery and family recognition that are modulated by the agent's structural traits. \textit{Intention} integrates belief, desire, and memory to produce concrete behavioral commitments; for technology adoption, agents evaluate their SES, prior experience, social influence, and family recommendation.

\subsubsection{Social Cognitive Memory (SCM)}
As agents learn through repeated interactions, they adapt their cognitive states by integrating new experiences with stored memories. Therefore, we design the SCM module to model this adaptive learning process. Specifically, we define its cognitive update process as
$$
(C_{\text{new}},\; M_{\text{new}}) \gets \operatorname{SCM}\bigl(\,f,\; Act,\; O,\; C_{\text{old}},\; M_{\text{old}}\,\bigr),
$$
where $C_{\text{new}}$ denotes the updated cognitive state, $M_{\text{new}}$ represents the updated memory structure, $Act$ is the action previously executed by the agent (\textit{i.e.}, the output of the BDI module), $O$ is the observed outcome, $C_{\text{old}}$ and $M_{\text{old}}$ are the prior states, and $f$ is the same LLM used in the BDI module.

\noindent \textbf{Multi-layer Trust Initialization.} Agents maintain \textit{specific} and \textit{general trust}. Specific trust is initialized based on social distance. Higher for the same group, moderate for adjacent groups, and lower for distant groups, with small random noise. During the simulation, trust levels are updated according to interaction outcomes. Positive interactions such as promise keeping or providing help increase trust, while negative interactions such as default or deception decrease trust, ensuring that trust values always evolve within a meaningful range.

\noindent \textbf{Norm Formation and Decay.} Normative cognition comprises \textit{general} and \textit{specific} dimensions. Initial norm levels correlate with SES. To capture the socially reinforced nature of norm maintenance, our system checks each agent’s reciprocity behavior every round. If no reciprocity is observed over ten consecutive rounds, norm decay is triggered.

\noindent \textbf{Memory and Reflection.} We divide the agent’s memory into two distinct layers, \textit{interaction-history} records specific interaction events, participants, and outcomes; \textit{cognitive-state} stores the evolving trajectories of trust and norms. After each round, an LLM-driven reflection process analyzes these memories each round, adjusting the agent's state and strategies for continuous adaptation.

\subsection{Multi-agent Simulation Environment}
\subsubsection{Social Network and Capital Evolution}

We design the social network to co-evolve interactively with the accumulation of social capital, forming a synergistic feedback loop during the simulation. Initially, ties are biased toward SES‑similar agents. Network density, computed as $Density = \frac{2|E|}{|V|(|V|-1)},$ where $|E|$ is the number of social ties and $|V|$ is the number of agents, measures the overall connectedness of the society. During the simulation, agents accumulate \textit{bonding social capital} from strong‑tie interactions and \textit{bridging social capital} from weak‑tie or cross‑group interactions. The update mechanism of social capital takes into account an agent’s capital type preference. When the type of interaction an agent engages in matches its own social capital preference, the accumulation rate is significantly accelerated; otherwise, it accumulates at a baseline rate.

\subsubsection{Two-phase Round-based Decision}
We extend turn‑based simulations with a two‑stage \textit{Proposal–Execution} process, capturing the social progression of collective action from initiation to fulfillment. Each round follows this sequence.

\noindent \textbf{Proposal phase.} Randomly selected initiators propose collective actions or technology adoptions. When an agent $a_i$ proposes, each neighbor $a_j$ responds using LLM-based reasoning that incorporates $a_j$'s social attributes, especially its specific trust toward $a_i$ and its perception of general norms. This captures how social attributes shape an agent's initial inclination toward collective decisions.

\noindent \textbf{Execution phase.} After accepting a proposal, each participant $a_i$ privately decides whether to honor the commitment, again using LLM-based reasoning that considers its current trust, norms, and social context. The collective action succeeds if the proportion of committing participants exceeds a threshold $\tau$ (default $\tau = 0.5$), with
$$
\text{Success} = \mathbb{1}\left[\frac{|\text{Committers}|}{|\text{Participants}|} > \tau \right].
$$

\section{Experimental Setup}
\subsection{Task Formulation}
To evaluate the capabilities of {\bf \name{}} for Putnam’s Social Capital Theory, we conduct two progressive tasks that examine how agents embody the dynamic coupling of social capital in interactions:

\noindent $\bullet$ \textbf{Modeling.} Can LLM agents reproduce the core dimensions of the theory and maintain consistency in group‑level decision patterns with real older adults in scenario‑based decisions?

\noindent $\bullet$ \textbf{Applying.} How do the core dimensions of Putnam’s Social Capital Theory affect the adaptation difficulties of elderly groups in smart elderly care?

\subsection{Agent and Simulation Configuration}
We evaluate three LLMs to assess the generalizability of our framework. Qwen2.5-14B-Instruct \citep{bai2023} serves as the default agent engine, while additional results for GPT-4 \citep{openai2024} and GLM-4 \citep{glm2024} are provided in Appendix~\ref{app:consistency} (\textit{modeling task}) \& \ref{app:tech} (\textit{applying task}). We set temperature = 0.7, top-p = 0.9. All key experiments are repeated five times; results are reported as means and standard errors. We generate 200 agents as detailed in Appendix \ref{app:data} (\textit{data construction}) \& \ref{app:demographic} (\textit{demographic profiles and behavioral analysis}). For \textit{modeling task}, we randomly sample 20 agents stratified by SES into low-, mid-, and high-SES groups for multi-round collective action experiments, each running for 25 rounds. For \textit{applying task}, we use all 200 agents for smart elderly care simulations, also over 25 rounds.

\subsection{Human Benchmarking} \label{sec:human_benchmarking}
We conduct parallel experiments with 20 real older adults ($age \geq 60$) recruited online for the \textit{modeling task}. Quota sampling matches the SES proportions of the 20 CGSS subsample. Each participant completed the same eight scenario-based decisions as the LLM agents (see Appendix~\ref{app:human1} for details). Prior to task completion, participants provided informed consent and completed self‑report inventories on demographics, social trust, and norms, aligned with agent initialization. Results are presented in \S\ref{sec:alignment}, further analyzed in Appendix~\ref{app:human2}.

\section{Modeling Results: Theory Replication}
In this section, we examine whether {\bf \name{}} can reproduce Putnam's core mechanisms and align with human decisions. We present results from three complementary perspectives: \textit{macro-level pattern replication}, \textit{ablation study}, and \textit{human‑agent alignment validation}.

\begin{figure}[ht] 
\vspace{-0.8\baselineskip}
  \vskip 0.2in
  \begin{center}
    \centerline{\includegraphics[width=\columnwidth]{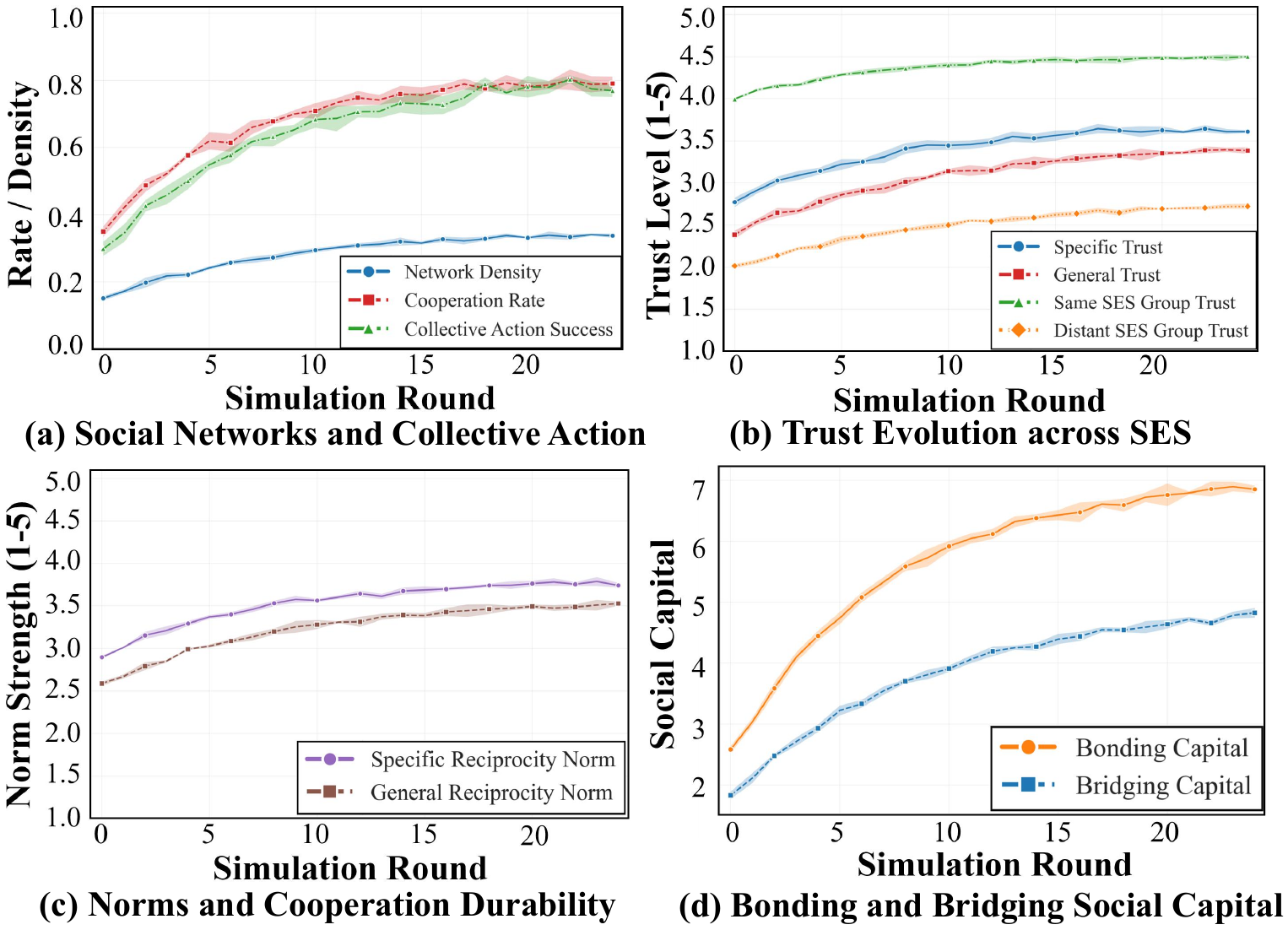}}
    \caption{
      Macro-level pattern replication of \textbf{(a) social network dimension}, \textbf{(b) trust dimension}, \textbf{(c) norms dimension}, and \textbf{(d) social capital accumulation dynamics} in Putnam's Social Capital Theory.
    }
    \label{Figure 4}
  \end{center}
  \vspace{-1.5\baselineskip}
\end{figure}

\subsection{Macro‑level Pattern Replication (Figure~\ref{Figure 4})}
To examine whether the simulated macro-level patterns align with Putnam’s theoretical predictions, we analyze the core dynamics across four aspects, with results shown in Figure~\ref{Figure 4}. Collective action outcomes are evaluated by the cooperation success rate, defined as the proportion of successful collective actions per round. In each round, agents interact within the social network and update their trust and norms based on historical experience.

\noindent $\bullet$ \textbf{Social Networks and Collective Action.} 

\textit{Figure~\ref{Figure 4}(a) demonstrates co‑evolution between network density, cooperation rate, and collective action success.} Over 25 rounds, network density increased from 0.158 to 0.342, while cooperation rate and success rate rose to 0.795 and 0.762, respectively. Correlation analysis shows strong positive associations between network density and both collective action success ($r \approx 0.976$, $p < 0.001$) and cooperation rate ($r \approx 0.993$, $p < 0.001$). Results are stable across five runs with low variance.

\begin{tcolorbox}[
    colback=gray!10,
    colframe=gray!10,
    boxrule=0.5pt,
    arc=2mm,
    left=5pt,
    right=5pt,
    top=3pt,
    bottom=3pt,
    fonttitle=\bfseries,
]
\ding{71} \textbf{Observation:} Initial network ties facilitate collective action, and the resulting cooperative success subsequently reinforces those ties, forming a self‑reinforcing dynamic.
\end{tcolorbox}

\begin{figure*}[ht]
  \vskip 0.2in
  \begin{center}
    \centerline{\includegraphics[width=\textwidth]{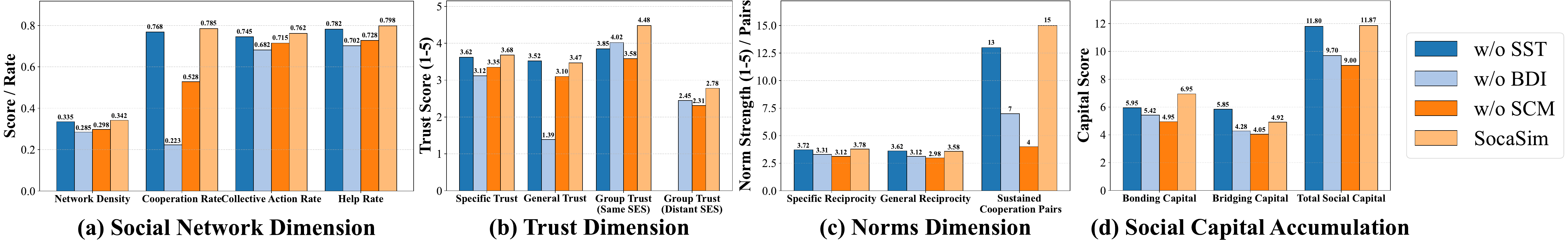}}
    \caption{
      Ablation study of SocaSim with SST, BDI, and SCM modules removed respectively.
    }
    \label{Figure 5}
  \end{center}
  \vspace{-1.5\baselineskip}
\end{figure*}

\begin{figure*}[ht]
 \vspace{-0.8\baselineskip}
  \vskip 0.2in
  \begin{center}
    \centerline{\includegraphics[width=\textwidth]{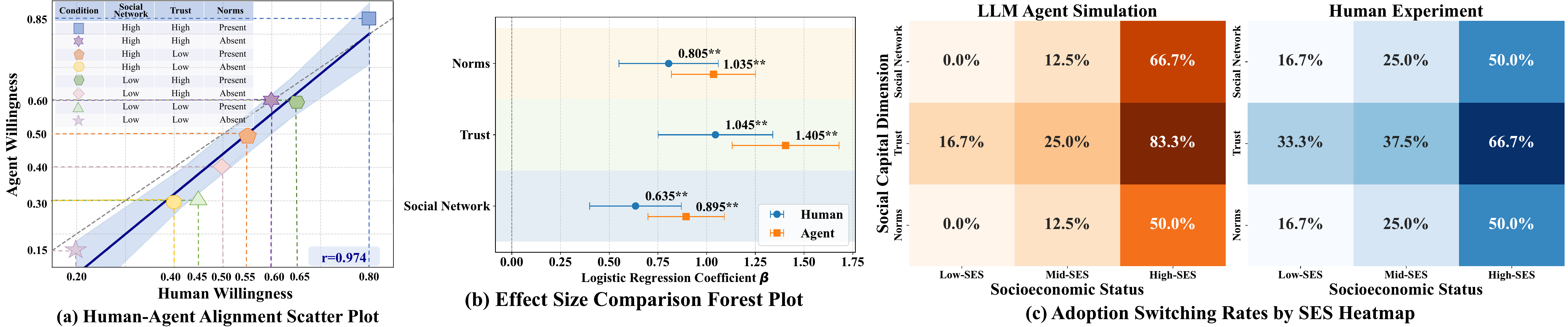}}
    \caption{
      Human-agent alignment: (a) scatter correlation, (b) coefficient comparison, (c) SES-switching heatmap.
    }
    \label{Figure 6}
  \end{center}
  \vspace{-2\baselineskip}
\end{figure*}

\noindent $\bullet$ \textbf{Trust Evolution across SES Groups.}

\textit{Figure~\ref{Figure 4}(b) shows the evolution of trust across different SES groups and trust types over 25 rounds.} Trust within the same SES group consistently remains the highest, rising from 3.98 to 4.48, while trust toward distant SES groups steadily increases from 2.02 to 2.78. Both specific and general trust grow in parallel, from 2.82 to 3.68 and from 2.46 to 3.47, respectively. These upward trends are consistent across runs with narrow variance.

\begin{tcolorbox}[
    colback=gray!10,
    colframe=gray!10,
    boxrule=0.5pt,
    arc=2mm,
    left=5pt,
    right=5pt,
    top=3pt,
    bottom=3pt,
    fonttitle=\bfseries,
]
\ding{71} \textbf{Observation:} Trust accumulates through interaction; trust toward distant SES groups grows more in absolute terms, partially bridging the gap while the hierarchy persists.
\end{tcolorbox}

\noindent $\bullet$ \textbf{Norms and Cooperation Durability.}

\textit{Figure~\ref{Figure 4}(c) illustrates the evolution of specific and general reciprocity norms over 25 rounds.} Specific reciprocity norms rise from 2.92 to 3.78, and general reciprocity from 2.61 to 3.58, maintaining a parallel growth trajectory. The simulation produces 15 sustained cooperation pairs.

\begin{tcolorbox}[
    colback=gray!10,
    colframe=gray!10,
    boxrule=0.5pt,
    arc=2mm,
    left=5pt,
    right=5pt,
    top=3pt,
    bottom=3pt,
    fonttitle=\bfseries,
]
\ding{71} \textbf{Observation:} Repeated interaction over time enables agents to internalize reciprocity norms, and once established, these norms sustain long‑term cooperation within the group even in the absence of immediate returns.
\end{tcolorbox}

\noindent $\bullet$ \textbf{Bonding and Bridging Social Capital.}

\textit{Figure~\ref{Figure 4}(d) compares the growth of bonding and bridging social capital over 25 rounds.} Bonding capital increases from 2.78 to 6.95, whereas bridging capital grows from 1.91 to 4.92, resulting in a final gap of 2.03. We further quantify cross‑SES connections through the economic connectivity (EC) index, defined as the proportion of high‑SES friends in a low‑SES agent's network, which rises steadily across rounds.

\begin{tcolorbox}[
    colback=gray!10,
    colframe=gray!10,
    boxrule=0.5pt,
    arc=2mm,
    left=5pt,
    right=5pt,
    top=3pt,
    bottom=3pt,
    fonttitle=\bfseries,
]
\ding{71} \textbf{Observation:} Bonding capital accumulates faster and reaches noticeably higher levels than bridging capital, while the steady rise in the EC index over time further suggests a gradual formation of cross‑SES weak ties.
\end{tcolorbox}

\begin{figure*}[ht] 
  \vspace{-1\baselineskip}  
    \vskip 0.2in  
    \begin{center}    
      \centerline{\includegraphics[width=\textwidth]{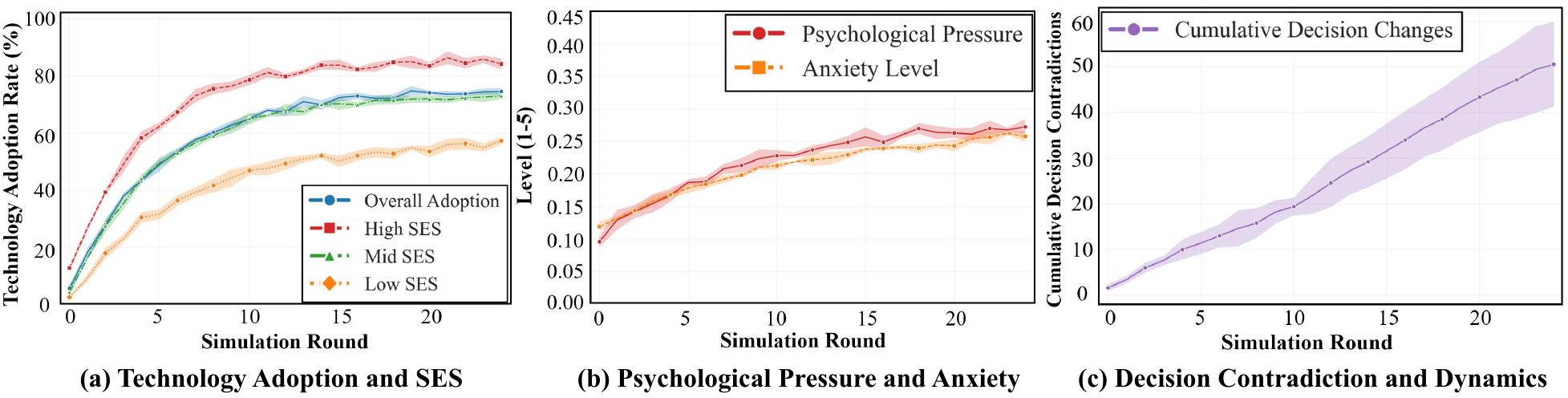}}   
        \caption{      
        Experimental simulation of social capital in smart elderly care technology adoption.  
        }    
      \label{Figure 7}  
    \end{center}  
  \vspace{-2\baselineskip}
\end{figure*}

\subsection{Ablation Study (Figure~\ref{Figure 5})}
To investigate how each component contributes to the simulation performance of SocaSim, we conduct an ablation study to validate the necessity of each module, with results shown in Figure~\ref{Figure 5}.

\noindent $\bullet$ \textbf{w/o SST.} 
Overall averages change only slightly, but the structural patterns are largely removed. In particular, the distinction between same‑SES and distant groups collapses, with bonding/bridging capital shifting from 6.95/4.92 to 5.95/5.85.
\begin{tcolorbox}[
    colback=gray!10,
    colframe=gray!10,
    boxrule=0.5pt,
    arc=2mm,
    left=5pt,
    right=5pt,
    top=3pt,
    bottom=3pt,
    fonttitle=\bfseries,
]
\ding{71} \textbf{Observation:} SST’s core function is to shape inter‑group structural differences, not to raise overall averages. The slight metric gains under w/o SST are due to the flattening of group disparities, not substantive improvement.
\end{tcolorbox}

\noindent $\bullet$ \textbf{w/o BDI.} 
Cooperation rate drops by 71.6\%, and general trust falls from 3.47 to 1.39.
\begin{tcolorbox}[colback=gray!10, colframe=gray!10, boxrule=0.5pt, arc=2mm, left=5pt, right=5pt, top=3pt, bottom=3pt]
\ding{71} \textbf{Observation:} BDI is the decision‑making core. Without it, agents cannot understand interaction contexts, and cooperation collapses.
\end{tcolorbox}

\noindent $\bullet$ \textbf{w/o SCM.} 
Sustained cooperative pairs decrease by 73.3\%, and specific reciprocity drops by 17.5\%.
\begin{tcolorbox}[colback=gray!10, colframe=gray!10, boxrule=0.5pt, arc=2mm, left=5pt, right=5pt, top=3pt, bottom=3pt]
\ding{71} \textbf{Observation:} SCM is the learning foundation. Without it, norms fail to internalize, and long‑term cooperation cannot be sustained.
\end{tcolorbox}

\subsection{Human Validation (Figure~\ref{Figure 6})} \label{sec:alignment}
To validate group-level consistency between LLM agents and real older adults, we compare their adoption willingness across eight scenarios that vary \textit{social network density (high/low)}, \textit{trust level (high/low)}, and \textit{reciprocity norm (present/absent)}. The 20 real older adults were recruited as described in \S~\ref{sec:human_benchmarking}. The results are summarized in Figure~\ref{Figure 6}.

\noindent $\bullet$ \textbf{Overall alignment (Figure~\ref{Figure 6}(a)).} \textit{Human and agent willingness rates are highly correlated (Pearson r = 0.974)}, with points tightly clustered around the identity line. Agents show slightly more extreme responses, higher under favorable conditions and lower under unfavorable ones.

\noindent $\bullet$ \textbf{Effect comparison (Figure~\ref{Figure 6}(b)).} All coefficients are positive and significant ($p < 0.01$), and the ranking is identical for both groups, \textit{trust has the strongest effect, followed by reciprocity norm, and then social network density}. Agents consistently produce larger coefficients, reflecting their cleaner causal reasoning that is less affected by noise such as emotion, fatigue, or social desirability, and thus showing more extreme responses.

\noindent $\bullet$ \textbf{Turnaround by SES (Figure~\ref{Figure 6}(c)).} For each dimension, we measure the switch from non-adoption to adoption when it improves, while keeping the other two dimensions fixed, and stratify by SES. In both groups, switching rates rise with SES and trust dominates. For low-SES agents, the trust switching rate is 16.7\% compared to 33.3\% for humans, while the rates for social network and norms are both 0\% versus 16.7\% for humans. For high-SES agents, the trust switching rate reaches 83.3\% compared to 66.7\% for humans. These results indicate that \textit{LLM agents may respond more strongly than humans to SES-based structural inequality}.

\begin{tcolorbox}[
    colback=gray!10,
    colframe=gray!10,
    boxrule=0.5pt,
    arc=2mm,
    left=5pt,
    right=5pt,
    top=3pt,
    bottom=3pt,
    fonttitle=\bfseries,
]
\ding{71} \textbf{Observation:} LLM agents closely reproduce the decision patterns of real older adults at the group level. Their responses provide clearer signals for identifying causal mechanisms of social capital, supporting external validity of our later applications in smart elderly care.
\end{tcolorbox}

\section{Applying Results: Elderly Care Study}

In this section, we investigate how the three core dimensions of Putnam’s Social Capital Theory affect the adaptation challenges of elderly groups in smart elderly care. We present results from two complementary perspectives: \textit{scenario simulation} and \textit{counterfactual intervention}.

\subsection{Scenario Simulation (Figure~\ref{Figure 7})}
We model the adoption of a smart elderly care platform as a binary choice using all 200 agents. The simulation tracks three multidimensional metrics: \textit{technology adoption rate}, defined as the proportion of agents that adopt the platform; \textit{psychological distress}, measured via psychological pressure value and anxiety level; and \textit{decision contradiction}, reflected by the frequency of adoption status switches. The results are shown in Figure~\ref{Figure 7}.

\noindent $\bullet$ \textbf{Technology Adoption and SES.} 

\textit{Figure~\ref{Figure 7}(a) shows adoption rates across SES groups over 25 rounds.} The high‑SES group rises steadily from 12.2\% to 84.5\% with minimal fluctuation. The low‑SES group increases slowly from 2.3\% to 56.8\%, with greater variability and wider confidence bands. The mid‑SES group stabilizes in the low‑70\% range. All groups grow rapidly in the first 10 rounds, then slow with diminishing returns.

\begin{tcolorbox}[
    colback=gray!10,
    colframe=gray!10,
    boxrule=0.5pt,
    arc=2mm,
    left=5pt,
    right=5pt,
    top=3pt,
    bottom=3pt,
    fonttitle=\bfseries,
]
\ding{71} \textbf{Observation:} Technology adoption rates are positively associated with SES, with high‑SES agents reaching higher levels and showing more stable trajectories than low‑SES agents.
\end{tcolorbox}

\noindent $\bullet$ \textbf{Psychological Pressure and Anxiety.}

\textit{Figure~\ref{Figure 7}(b) illustrates the evolution of psychological pressure and anxiety.} Both metrics rise steeply in early rounds (pressure from 0.10 to 0.27, anxiety from 0.12 to 0.26) and then stabilize. Pressure remains slightly higher than anxiety throughout, and variance widens modestly over time.

\begin{tcolorbox}[
    colback=gray!10,
    colframe=gray!10,
    boxrule=0.5pt,
    arc=2mm,
    left=5pt,
    right=5pt,
    top=3pt,
    bottom=3pt,
    fonttitle=\bfseries,
]
\ding{71} \textbf{Observation:} Psychological pressure and anxiety increase rapidly during early adoption rounds, clearly indicating that technology adoption induces significant stress that later stabilizes but leaves an elevated baseline level.
\end{tcolorbox}

\noindent $\bullet$ \textbf{Decision Contradiction and Dynamics.}

\textit{Figure~\ref{Figure 7}(c) presents the evolution of cumulative decision contradictions.} The metric rises steadily from near 0 to about 50 by round 25. Its growth rate is consistent without clear inflection points. Variance across five runs diverges after round 10.

\begin{tcolorbox}[
    colback=gray!10,
    colframe=gray!10,
    boxrule=0.5pt,
    arc=2mm,
    left=5pt,
    right=5pt,
    top=3pt,
    bottom=3pt,
    fonttitle=\bfseries,
]
\ding{71} \textbf{Observation:} Decision contradictions accumulate linearly over time, with increasing variance across runs, suggesting persistent decision uncertainty and growing individual differences.
\end{tcolorbox}

\subsection{Counterfactual Intervention (Table~\ref{Table 1})}

To enhance causal interpretability and simulate a policy intervention, we design a counterfactual experiment. Since trust is the strongest dimension (Figure~\ref{Figure 6}(b)), we increase the initial trust of low‑SES agents by 1.0 (capped at 5.0) while keeping all other parameters unchanged, mimicking a targeted policy for disadvantaged groups.

\begin{figure}[ht]
\vspace{-1.5\baselineskip}  
  \vskip 0.2in
  \begin{center}
    \centerline{\includegraphics[width=\columnwidth]{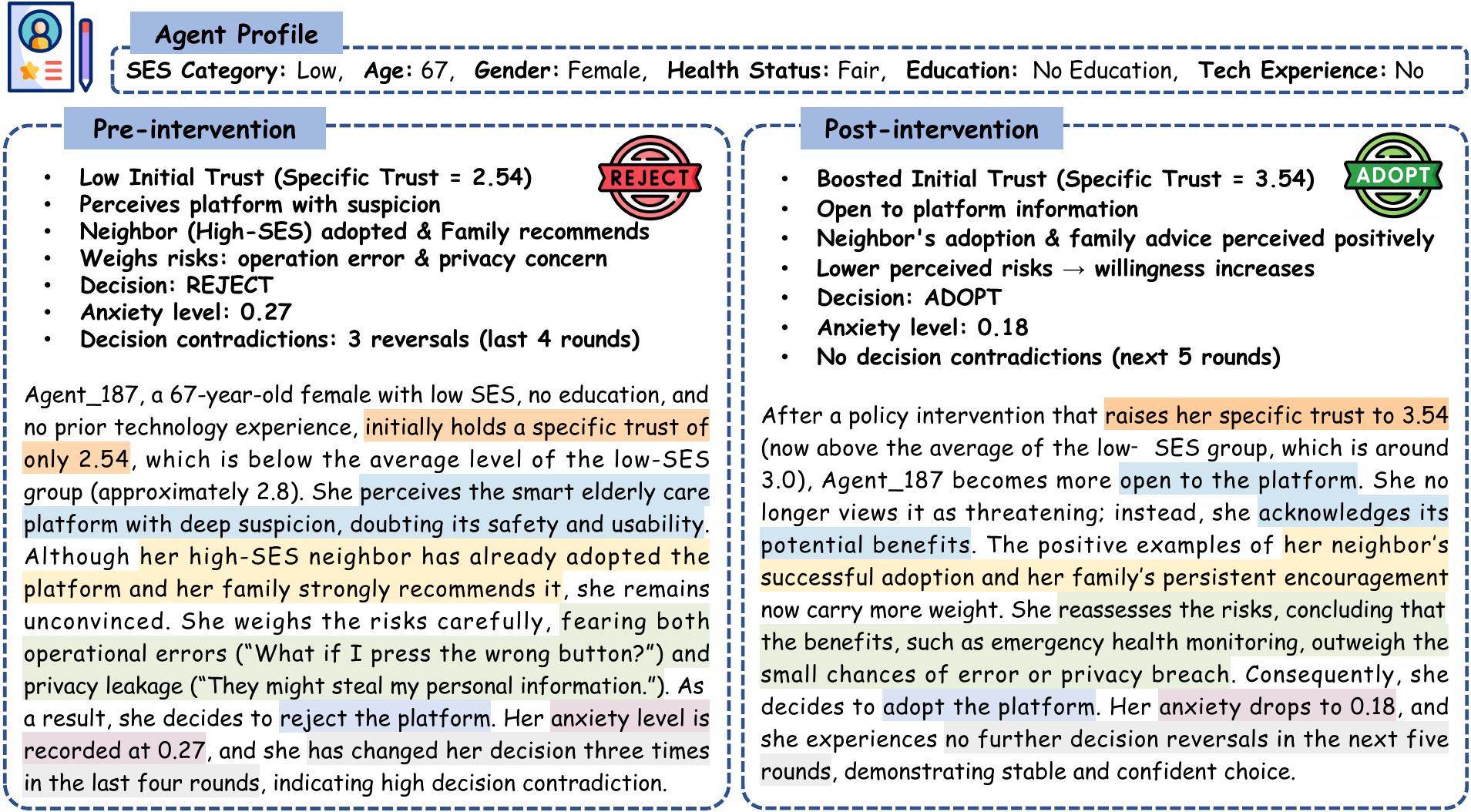}}
    \caption{
      Case study: low-SES trust intervention.
    }
    \label{Figure 8}
  \end{center}
  \vspace{-2.5\baselineskip}
\end{figure}

\begin{table}[ht]
\centering
\resizebox{\columnwidth}{!}{%
\small
\begin{tabular}{c c c c}
    \hline
    \textbf{Metric} & \textbf{Pre-intervention} & \textbf{Post-intervention} & \textbf{Relative change} \\
    \hline
    Technology Adoption Rate  & 0.732 {\scriptsize $\pm$0.028} & 0.845 {\scriptsize $\pm$0.024} & +15.4\% \\
    Psychological Pressure    & 0.268 {\scriptsize $\pm$0.032} & 0.215 {\scriptsize $\pm$0.028} & -19.8\% \\
    Anxiety Level             & 0.255 {\scriptsize $\pm$0.028} & 0.198 {\scriptsize $\pm$0.022} & -22.4\% \\
    Decision Contradictions   & 51 {\scriptsize $\pm$8}        & 38 {\scriptsize $\pm$6}        & -25.5\% \\
    \hline
\end{tabular}%
}
\caption{Overall performance of 200 agents at round 25 under factual and counterfactual conditions.}
\label{Table 1}
\vspace{-0.5\baselineskip}
\end{table}

We illustrate the effect using a case study (Figure~\ref{Figure 8}). Before intervention, low trust causes hesitation, repeated decision contradictions, and eventual rejection. After trust is raised, the agent adopts stably, with anxiety and contradictions dropping sharply. This demonstrates trust's causal role in stability. Table~\ref{Table 1} compares the aggregate outcomes at round 25. After the intervention, technology adoption rises by 15.4\%, pressure and anxiety fall by 19.8\% and 22.4\%, and decision contradictions decrease by 25.5\%. These results indicate that boosting initial trust of low‑SES groups reduces the digital divide and improves decision stability.

\section{Discussion}

Our findings carry several important implications for policy, highlight the unique strengths and limitations of using LLM agents for social simulation, and open up new avenues for interdisciplinary research. For a more detailed discussion of these points, we refer the readers to Appendix~\ref{app:discussion}.

\section{Related Work}
\textbf{Putnam’s Social Capital Theory.} Putnam's Social Capital Theory defines social capital as rooted in social network, trust, and norms \citep{putnam1993, putnam2000}. It serves as a core paradigm for explaining civic engagement, government performance, and democratic functioning \citep{tzanakis2013, dodd2015}, and is widely applied to examine regional development \citep{annamalah2023}, public health \citep{carpiano2006}, and education attainment \citep{goddard2003, careemdeen2021} as outcomes of resource accumulation within community structures. Recent studies further explore its role in crises \citep{chawa2023} and sustainable development \citep{sumi2025}. However, conventional empirical methods face practical constraints on experimental control and replication.

\noindent \textbf{LLM-based Multi-agent Social Simulation.} The rise of LLMs has enabled human-like agents capable of reasoning and social interaction \citep{wang2026,sudhakar2025,wang20251}. These agents perform complex social tasks \citep{zhang20251,xu2025}, leading to growing use of LLM-based social simulation across scenarios \citep{piao2025,yang2025,zhang2025}. For example, generative agents \citep{piao2025} simulate intricate social behaviors; ProSIM \citep{zhou2025} models prosocial behavior; and SOTOPIA-$\Omega$ \citep{zhang2025} integrates human negotiation strategies. However, existing frameworks have not yet provided a reproducible and controllable environment for systematically instantiating Putnam's Social Capital Theory, nor have they offered process-level interpretability of how trust accumulates or norms are internalized. Our work aims to address this gap.

\section{Conclusion}
We presented {\bf \name{}}, a novel LLM‑based multi‑agent simulation framework to model and apply Putnam's Social Capital Theory. By integrating social network evolution, trust dynamics, and norm propagation into a unified environment, we analyzed collective‑action behaviors and extended the framework to smart elderly care. Our results demonstrate a consistent alignment between agent behaviors and human social dynamics, revealing how social capital shapes group coordination and technology adoption. Moreover, {\bf \name{}} offers a scalable, reproducible approach to investigate social mechanisms across social science domains.

\section*{Limitations}

In this section, we discuss the limitations of our work as follow:

\begin{itemize}
    \item Our current framework models the dynamic evolution of social capital primarily through text-based interactions, whereas in reality, trust building and norm propagation often rely on multimodal signals such as speech and images. Incorporating multimodal information is an important direction to enhance simulation realism, but it also brings challenges in data acquisition, cross-modal alignment, and ethics. Therefore, exploring multimodal social interaction mechanisms is a highly valuable research direction for future work.
    \item Due to recruitment costs and resource constraints, the number of participants in our human alignment experiment is limited, which may affect the statistical generalizability of the findings. Future work should increase the sample size and include more diverse populations to more robustly analyze alignment between simulated and real behaviors across different demographic dimensions.
    \item Our study is based on CGSS data and has not incorporated comparative analyses across diverse cultural contexts. The accumulation and dynamics of social capital may be shaped by cultural values, institutional settings, and related factors. Cross-cultural comparisons are thus a promising direction to evaluate both generalizability and context-specificity.
\end{itemize}

\section*{Ethics Statement}

As the use of LLMs for social simulation grows, it is crucial to consider the ethical implications of deploying such systems to model social capital dynamics and elderly care behaviors. While this work explores the potential of LLM‑based multi-agent simulations to study Putnam’s Social Capital Theory, all experiments are conducted solely within controlled research settings and remain an early-stage theoretical exploration.

The {\bf \name{}} framework and all simulations are designed for scientific theory validation and methodological development rather than for direct real‑world decision making, policy formulation, or replacing the complexity of human social systems. We strongly emphasize that any potential application of our findings to smart elderly care policy, community governance, or service design must involve human expert oversight, ethical review, and careful consideration of fairness to mitigate risks arising from model simplifications or data biases.

All demographic data used for agent initialization are drawn from the 2023 CGSS elderly subsample, which is released as publicly available aggregate statistics and contains no personally identifiable information. The construction of synthetic agents does not involve direct simulation or identity mapping of real individuals. For our human‑agent alignment validation, 20 real older volunteers participated with informed consent. The study consisted solely of anonymous questionnaires and scenario-based decision tasks, with no interventional or deceptive procedures. Participants’ personal information was used only for SES stratification and anonymized analysis, and they received reasonable compensation in accordance with local ethical guidelines. The research protocol was approved by the institutional ethics board.

Looking forward, we will continue to focus on fairness, transparency, and accountability in agent-based social simulation. We will actively expand cross-cultural and multimodal simulation capabilities, and develop human-in-the-loop experimental paradigms. We remain committed to advancing this field toward responsible, inclusive, and socially beneficial directions.


\bibliography{main}

\begin{figure*}[ht]
  \vskip 0.2in
  \begin{center}
    \centerline{\includegraphics[width=\textwidth]{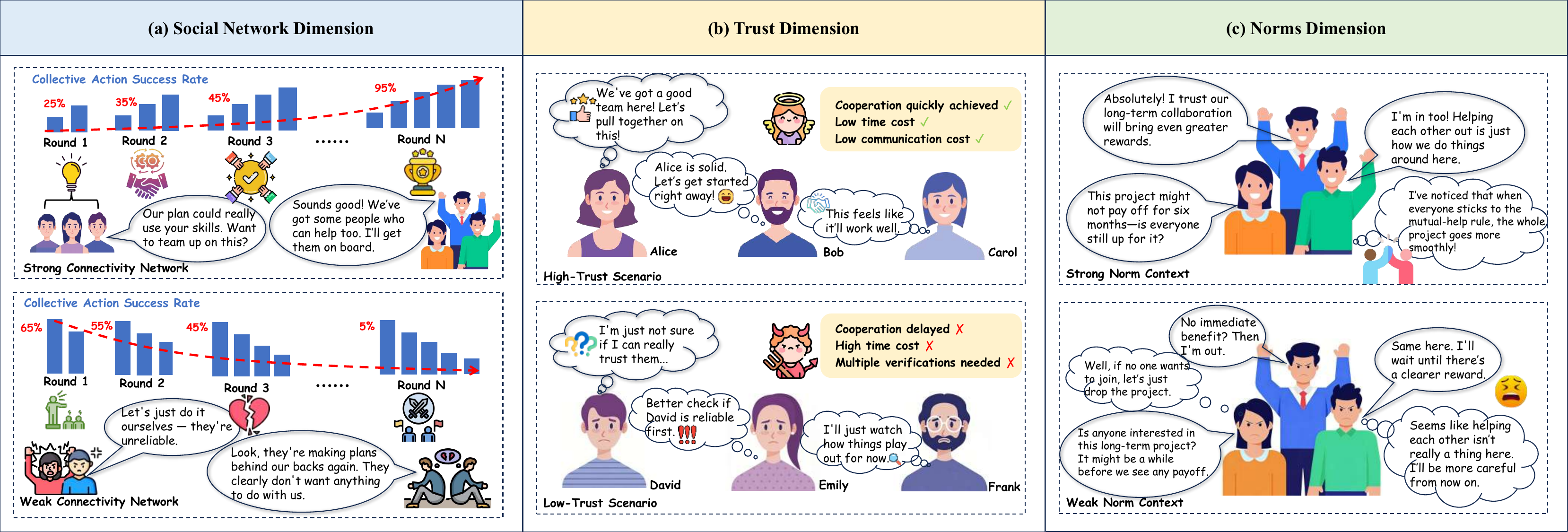}}
    \caption{Illustration and examples of three core dimensions in Putnam’s Social Capital Theory.}
    \label{Figure 2}
  \end{center}
  \vspace{-1.5\baselineskip}
\end{figure*}

\newpage
\appendix

\section{Supplementary Preliminary} \label{app:preliminary}

Putnam’s Social Capital Theory \citep{putnam1993,putnam2000} offers a theoretical blueprint of how social network, trust, and norms interact in self‑reinforcing ways \citep{Ostrom2000,Sonderskov2016}, and it has influenced work in fields ranging from political science and public administration to economic development \citep{mikiewicz2021,zhou2023,thang2025}. This section further explains these three dimensions, as illustrated in Figure~\ref{Figure 2}, by showing how they jointly shape a group’s ability to overcome collective‑action dilemmas. We also provide concrete positive and negative examples for each dimension, and highlight how the interplay among these dimensions affects coordination, responses to shared problems, and sustained participation.

\subsection*{1. Social Network Dimension (Figure~\ref{Figure 2}(a))}

Putnam highlights the importance of civic and associational networks in enabling coordination and mutual support \citep{putnam1993}. In this sense, the social network dimension concerns the existence, density, and stability of social ties, which determine how easily information, resources, and assistance can circulate within a group.

\paragraph{Positive example (strong connectivity network).}
In a community where people know each other well and interact frequently, a proposal for a joint project spreads quickly through trusted channels. Members readily accept, offer help, and recruit collaborators. As a result, coordination is smooth, and the project succeeds with minimal delays.

\paragraph{Negative example (weak connectivity network).}
In a fragmented community where residents rarely interact, a proposal for a joint project is met with suspicion. Some members prefer to act alone, believing others are unreliable, while others assume their neighbors are plotting behind their backs. Consequently, cooperation fails, the project stalls, and potential benefits are lost.

\subsection*{2. Trust Dimension (Figure~\ref{Figure 2}(b))}

Trust is another central component of social capital. Putnam argues that trust facilitates coordination by reducing uncertainty about others’ behavior and lowering the costs of collective action \citep{putnam1993}. When trust is present, individuals are more willing to rely on one another and to participate without monitoring or verification.

\paragraph{Positive example (high-trust scenario).}
In a group where members have consistently kept their promises, a new collaborative task is met with immediate and full commitment. People assume others will contribute fairly, so they freely invest effort without hesitation, openly share information, and complete the task efficiently.

\paragraph{Negative example (low-trust scenario).}
In a group where past experiences have been marked by broken commitments, even a well‑designed proposal is met with suspicion. Members insist on checking each other’s reliability, avoid committing, and prefer to watch how things unfold. They spend excessive time verifying intentions and drafting detailed contracts. As a result, cooperation is delayed, costs rise, and the goals are often not reached.

\subsection*{3. Norms Dimension (Figure~\ref{Figure 2}(c))}

Putnam also emphasizes the importance of shared norms, especially norms of reciprocity \citep{putnam2000}. This dimension concerns widely accepted expectations about appropriate behavior, such as mutual support, reciprocal help, and continued participation. Strong norms make sustained collective action more likely in situations where immediate returns are uncertain or delayed.

\paragraph{Positive example (strong norm context).}
In a community where mutual aid is taken for granted, a long‑term infrastructure project that will only pay off after many months receives broad support from the start. Members openly affirm that helping each other is how things are done, and they trust that sticking to the mutual‑help rule will eventually benefit everyone. They contribute without expecting an immediate reward, and the project proceeds steadily to successful completion.

\paragraph{Negative example (weak norm context).}
In a community where people primarily look after their own short‑term interests, the same long‑term project meets immediate resistance. Each potential participant focuses on immediate personal gain and refuses to join without an instant payoff. Some even suggest abandoning the project altogether, while others prefer to wait until a clearer reward appears. As a result, the project never gets off the ground, and the prevailing attitude is that helping each other is not a shared expectation.

This theoretical logic directly supports our study in two respects. First, it provides a framework for formalizing how social capital shapes a group’s capacity to overcome collective-action dilemmas, which forms the basis of our theoretical \textit{modeling task}. Second, it offers an analytical lens for smart elderly care, where adaptation depends not only on individual capability, but also on the surrounding structure of social relationships, trust, and supportive norms. This motivates our \textit{applying task}, in which we examine how the three dimensions of social capital are related to the adaptation dilemmas faced by older adults in smart elderly care.

\begin{table*}[htbp]
  \centering
  
  \begin{tabular*}{\textwidth}{@{\extracolsep{\fill}} c c c c c @{}}
    \hline
    \textbf{LLM Model} & \textbf{Network Density} & \textbf{Cooperation Rate} & \textbf{Collective Action Rate} & \textbf{Help Rate} \\
    \hline
    Qwen2.5-14B & 0.342 {\scriptsize $\pm$0.028} & 0.795 {\scriptsize $\pm$0.035} & 0.762 {\scriptsize $\pm$0.042} & 0.798 {\scriptsize $\pm$0.031} \\
    GPT-4 & 0.325 {\scriptsize $\pm$0.032} & 0.752 {\scriptsize $\pm$0.041} & 0.728 {\scriptsize $\pm$0.048} & 0.765 {\scriptsize $\pm$0.038} \\
    GLM-4 & 0.308 {\scriptsize $\pm$0.035} & 0.718 {\scriptsize $\pm$0.045} & 0.695 {\scriptsize $\pm$0.052} & 0.732 {\scriptsize $\pm$0.042} \\
    \hline
  \end{tabular*}
  \caption{Results of social network dimension.}
  \label{table3}
  
  \vspace{0.5\baselineskip}

  \begin{tabular*}{\textwidth}{@{\extracolsep{\fill}} c c c c c @{}}
    \hline
    \textbf{LLM Model} & \textbf{Specific Trust} & \textbf{General Trust} & \shortstack[c]{\textbf{Group Trust}\\\textbf{(Same SES)}} & \shortstack[c]{\textbf{Group Trust}\\\textbf{(Distant SES)}} \\
    \hline
    Qwen2.5-14B & 3.68 {\scriptsize $\pm$0.12} & 3.47 {\scriptsize $\pm$0.14} & 4.48 {\scriptsize $\pm$0.08} & 2.78 {\scriptsize $\pm$0.11} \\
    GPT-4 & 3.55 {\scriptsize $\pm$0.15} & 3.38 {\scriptsize $\pm$0.16} & 4.35 {\scriptsize $\pm$0.10} & 2.65 {\scriptsize $\pm$0.13} \\
    GLM-4 & 3.42 {\scriptsize $\pm$0.18} & 3.25 {\scriptsize $\pm$0.19} & 4.22 {\scriptsize $\pm$0.12} & 2.52 {\scriptsize $\pm$0.15} \\
    \hline
  \end{tabular*}
  \caption{Results of trust dimension.}
  \label{table4}
  
  \vspace{0.5\baselineskip}
  
  \begin{tabular*}{\textwidth}{@{\extracolsep{\fill}} c c c c @{}}
    \hline
    \textbf{LLM Model} & \textbf{Specific Reciprocity} & \textbf{General Reciprocity} & \textbf{Sustained Cooperation Pairs} \\
    \hline
    Qwen2.5-14B & 3.78 {\scriptsize $\pm$0.11} & 3.58 {\scriptsize $\pm$0.13} & 15 {\scriptsize $\pm$2} \\
    GPT-4 & 3.62 {\scriptsize $\pm$0.14} & 3.42 {\scriptsize $\pm$0.15} & 13 {\scriptsize $\pm$3} \\
    GLM-4 & 3.48 {\scriptsize $\pm$0.16} & 3.28 {\scriptsize $\pm$0.18} & 11 {\scriptsize $\pm$3} \\
    \hline
  \end{tabular*}
  \caption{Results of norms dimension.}
  \label{table5}
  
  \vspace{0.5\baselineskip}
  
  \begin{tabular*}{\textwidth}{@{\extracolsep{\fill}} c c c c @{}}
    \hline
    \textbf{LLM Model} & \textbf{Bonding Social Capital} & \textbf{Bridging Social Capital} & \textbf{Total Social Capital} \\
    \hline
    Qwen2.5-14B & 6.95 {\scriptsize $\pm$0.42} & 4.92 {\scriptsize $\pm$0.35} & 11.87 {\scriptsize $\pm$0.68} \\
    GPT-4 & 6.58 {\scriptsize $\pm$0.48} & 4.65 {\scriptsize $\pm$0.40} & 11.23 {\scriptsize $\pm$0.75} \\
    GLM-4 & 6.22 {\scriptsize $\pm$0.52} & 4.38 {\scriptsize $\pm$0.45} & 10.60 {\scriptsize $\pm$0.82} \\
    \hline
  \end{tabular*}
  \caption{Results of social capital accumulation.}
  \label{table6}
  
\end{table*}

\begin{figure*}[!ht]
  \vskip 0.2in
  \begin{center}
    \centerline{\includegraphics[width=\textwidth]{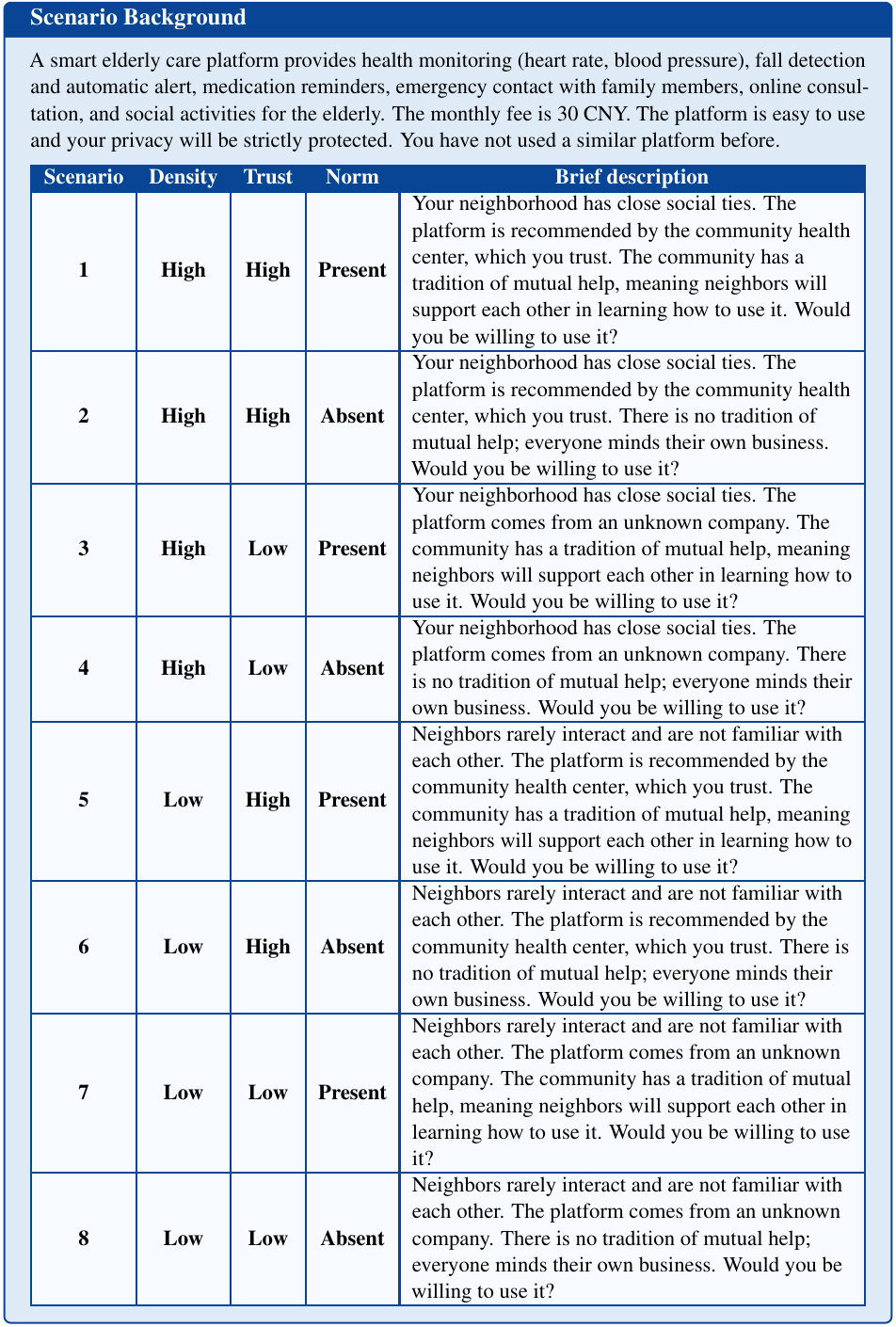}}
    \caption{
     Eight scenarios with full factorial design of social network density, trust level, and reciprocity norm.
    }
    \label{Figure 12}
  \end{center}
  \vspace{-2\baselineskip}
\end{figure*}

\section{Cross‑Model Consistency in Macro‑level Pattern Replication} \label{app:consistency}
To address reproducibility concerns and validate the generalizability of our findings, we conducted additional experiments using three LLMs with distinct architectures: Qwen‑2.5‑14B\citep{bai2023}, GPT‑4\citep{openai2024}, and GLM‑4\citep{glm2024}. The results demonstrate consistent replication of the macro‑level patterns of Putnam's Social Capital Theory across its three core dimensions (social network, trust, and norms) and two forms of social capital (bonding and bridging), based on data from the final round of simulation. All models reproduce core patterns aligned with Putnam's Social Capital Theory.

\paragraph{ $\bullet$ Social Network Dimension (Table~\ref{table3}):} Every model exhibits high network density, cooperation rate, collective action rate, and help rate, consistent with the macro‑level pattern that dense networks facilitate cooperation and mutual aid.

\paragraph{ $\bullet$ Trust Dimension (Table~\ref{table4}):} The results show that specific trust, general trust, and group trust (same SES) are higher than group trust (distant SES), replicating the structural nature of trust.

\paragraph{ $\bullet$ Norms Dimension (Table~\ref{table5}):} High rates of specific and general reciprocity, along with sustained cooperation pairs, indicate that norms are maintained and reinforced through long‑term interaction, reproducing the macro‑level patterns.

\paragraph{ $\bullet$ Social Capital Accumulation (Table~\ref{table6}):} All models exhibit growth in both bonding and bridging capital by the end of the simulation, confirming the dynamic accumulation of social capital as a macro‑level phenomenon.

All reported metrics are averaged over five independent runs with standard deviations shown. The consistent trends across models underscore the robustness and generalizability of our macro‑level pattern replication.

\section{Human Experiment Details} \label{app:human}
\subsection{Experimental Procedure} \label{app:human1}
We recruited 20 real older adults online to participate in the modeling task. Given the digital literacy barriers among some older adults, the experiment was designed as a simple questionnaire rather than a complex online platform. Each participant was presented with the same fixed scenario background followed by eight scenario‑based decisions in random order. For each scenario, participants were asked to choose between two options: \textit{willing} or \textit{unwilling}. The questionnaire also collected demographic information, self‑reported social trust, and reciprocity norms. The fixed scenario background and the eight scenarios are illustrated in Figure~\ref{Figure 12}.

Prior to the experiment, all participants provided written informed consent and were clearly informed about the purpose of the study, the voluntary nature of their participation, and the measures taken to protect their privacy and data. The entire session took approximately 15 minutes to complete, and each participant received 20 CNY, equivalent to 80 CNY per hour, which is comparable to the typical hourly rate for simple online tasks in China.

\begin{figure*}[p]  
  \centering
  \includegraphics[width=\textwidth]{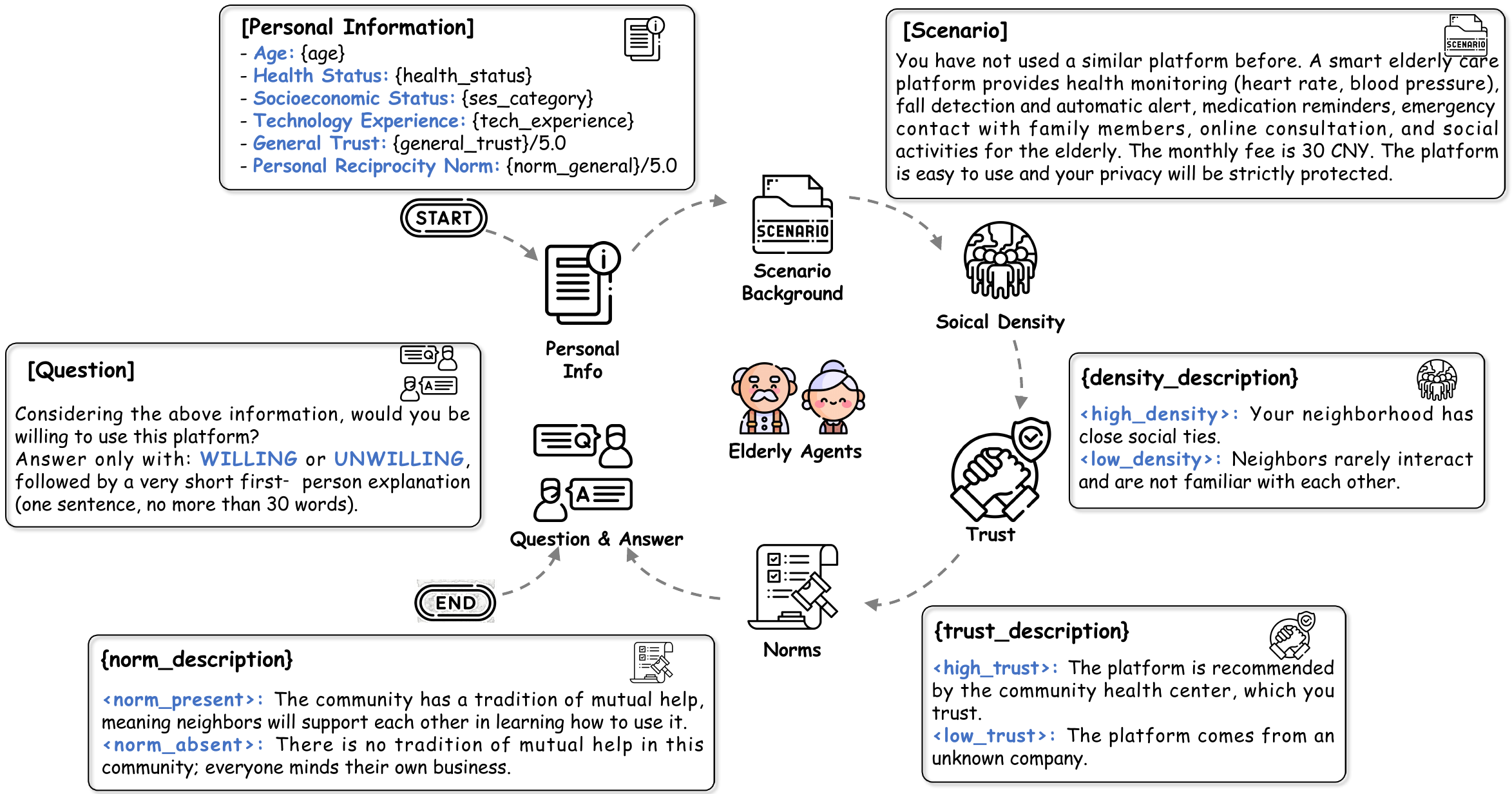}
  \caption{
    Prompt structure and the factorial design of social density, trust, and norms across eight scenarios.
  }
  \label{Figure 11}

  \vspace{0.5cm}
  
  \small
  \setlength{\arrayrulewidth}{1pt}
  \renewcommand{\arraystretch}{1.15}
  \setlength{\tabcolsep}{3.5pt}
  \rowcolors{3}{mycolor}{white}
  \begin{tabular}{cccccccccc}
    \hline
    \rowcolor{myblue}
    \textcolor{white}{\textbf{Agent ID}} & \textcolor{white}{\textbf{Age}} & 
    \textcolor{white}{\textbf{Gender}} & \textcolor{white}{\textbf{Health}} & 
    \textcolor{white}{\textbf{Education}} & \textcolor{white}{\textbf{Income}} & 
    \textcolor{white}{\textbf{Occupation}} & \textcolor{white}{\textbf{Initial Trust}} & 
    \textcolor{white}{\textbf{Initial Norm}} & \textcolor{white}{\textbf{Tech Exp.}} \\
    \hline
    \multicolumn{10}{c}{\cellcolor{white}\textbf{Low-SES (6 agents)}} \\
    \hline
    $\text{Agent}_1$  & 72 & Female & Poor & No Education     & 800  & Unemployed    & 2.2 & 2.1 & No \\
    $\text{Agent}_2$  & 68 & Male   & Fair & Primary          & 1,200 & Farmer        & 2.5 & 2.4 & Yes \\
    $\text{Agent}_3$  & 75 & Female & Poor & No Education     & 600  & Unemployed    & 1.8 & 1.9 & No \\
    $\text{Agent}_4$  & 66 & Male   & Good & Primary          & 1,500 & Service       & 2.6 & 2.5 & Yes \\
    $\text{Agent}_5$  & 71 & Female & Fair & Primary          & 1,000 & Farmer        & 2.3 & 2.2 & No \\
    $\text{Agent}_6$  & 69 & Male   & Poor & No Education     & 900  & Unemployed    & 2.0 & 2.0 & No \\
    \hline
    \multicolumn{10}{c}{\cellcolor{white}\textbf{Mid-SES (8 agents)}} \\
    \hline
    $\text{Agent}_7$  & 65 & Female & Good     & High School      & 3,500 & Clerk         & 3.0 & 3.0 & Yes \\
    $\text{Agent}_8$  & 70 & Male   & Fair     & Secondary        & 2,800 & Service       & 2.8 & 2.9 & No \\
    $\text{Agent}_9$  & 63 & Female & Good     & High School      & 4,200 & Clerk         & 3.2 & 3.3 & Yes \\
    $\text{Agent}_{10}$ & 67 & Male   & Fair     & Secondary        & 3,200 & Technician    & 2.9 & 3.0 & No \\
    $\text{Agent}_{11}$ & 74 & Female & Fair     & High School      & 3,000 & Clerk         & 2.7 & 2.8 & No \\
    $\text{Agent}_{12}$ & 82 & Male   & Good     & College          & 5,000 & Professional  & 3.5 & 3.6 & Yes \\
    $\text{Agent}_{13}$ & 68 & Female & Fair     & Secondary        & 2,600 & Service       & 2.6 & 2.7 & No \\
    $\text{Agent}_{14}$ & 71 & Male   & Fair     & High School      & 3,800 & Technician    & 3.0 & 3.2 & No \\
    \hline
    \multicolumn{10}{c}{\cellcolor{white}\textbf{High-SES (6 agents)}} \\
    \hline
    $\text{Agent}_{15}$ & 64 & Male   & Good     & University       & 8,000 & Professional  & 4.0 & 4.0 & Yes \\
    $\text{Agent}_{16}$ & 69 & Female & Good     & College          & 7,500 & Manager       & 3.7 & 3.8 & Yes \\
    $\text{Agent}_{17}$ & 61 & Male   & Very Good& University       & 13,000 & Professional  & 4.3 & 4.2 & Yes \\
    $\text{Agent}_{18}$ & 73 & Male   & Good     & College          & 9,200 & Manager       & 3.8 & 3.9 & Yes \\
    $\text{Agent}_{19}$ & 66 & Female & Good     & University       & 10,500 & Professional  & 4.1 & 4.1 & Yes \\
    $\text{Agent}_{20}$ & 70 & Male   & Fair     & College          & 6,300 & Manager       & 3.1 & 3.6 & No \\
    \hline
  \end{tabular}
  \captionof{table}{
    Initial profiles of the 20 agents used in the modeling task for the macroscopic pattern replication and human-agent alignment experiments. All agents are drawn from the CGSS elderly subsample. SES groups are indicated by the row headings. Tech Exp.: Yes = Has prior experience, No = No prior experience.
  }
  \label{tab:agent-profiles}
\end{figure*}

\textbf{Each LLM agent received a prompt whose scenario, question, and response format were designed to be completely identical to those used in the human survey}. The community‑level social variables, namely social network density (high/low), trust level (high/low), and reciprocity norm (present/absent), were fully crossed in a factorial design, resulting in eight distinct scenarios. The presentation order of the eight scenarios was randomized for each agent. The prompt template and the eight community scenarios are illustrated in Figure~\ref{Figure 11}. In addition, Table~\ref{tab:agent-profiles} presents the 20 agents that were randomly drawn from the 200 elderly subsample for the modeling task, covering low-SES, mid-SES, and high-SES strata.

\subsection{Analysis: Divergence Between Simulated and Real older Adults} \label{app:human2}

\begin{figure*}[!ht]
\vspace{-0.5\baselineskip}
\centering
\includegraphics[width=\textwidth]{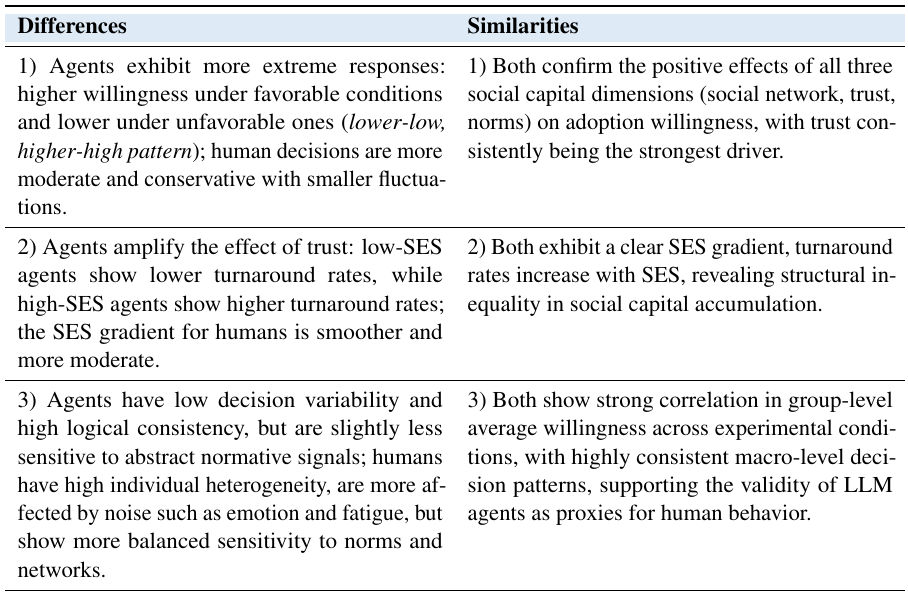}

\caption{Comparison of LLM Agents vs. Real Older Adults.}
\label{tab:comparison}

\end{figure*}

Based on the human alignment experiments in the modeling task, we further analyze the similarities and differences between real older adults and LLM agents, as summarized in Figure~\ref{tab:comparison}.

\begin{figure*}[!ht]
\centering
\includegraphics[width=\textwidth]{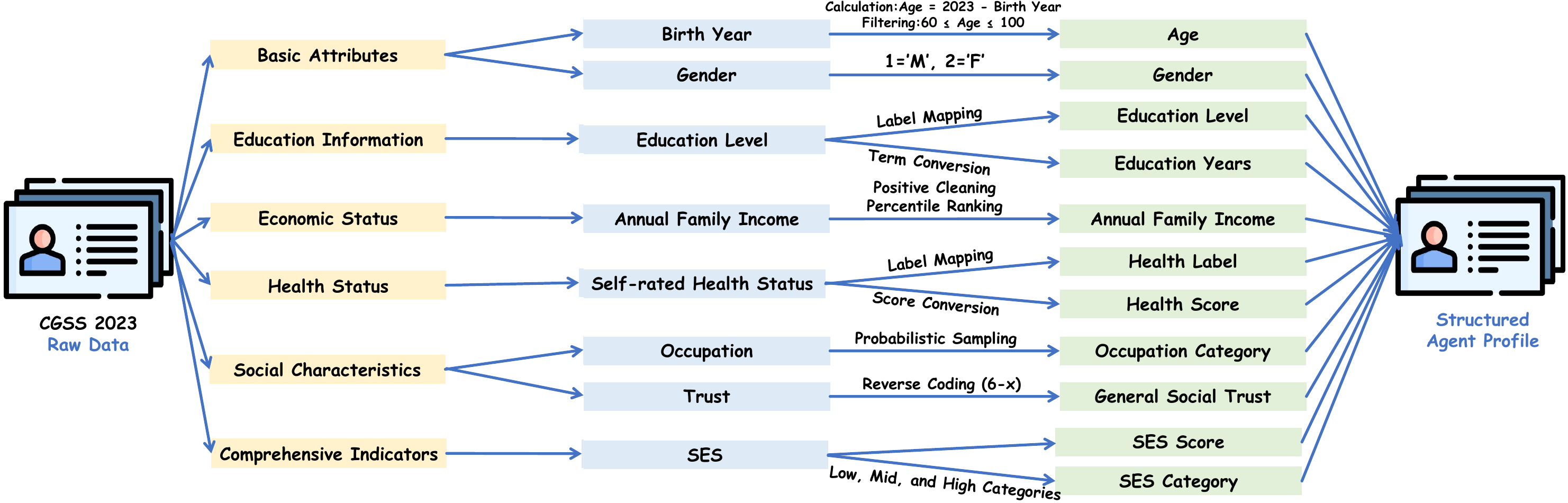}
\caption{Constructing Agent Profiles Through a Multi-Step Data Processing Pipeline of CGSS 2023 Dataset.}
\label{Figure 9}
\end{figure*}

\begin{figure*}[ht]
 \vspace{-0.8\baselineskip}
 \vskip 0.2in
 \begin{center}
   \centerline{\includegraphics[width=\textwidth]{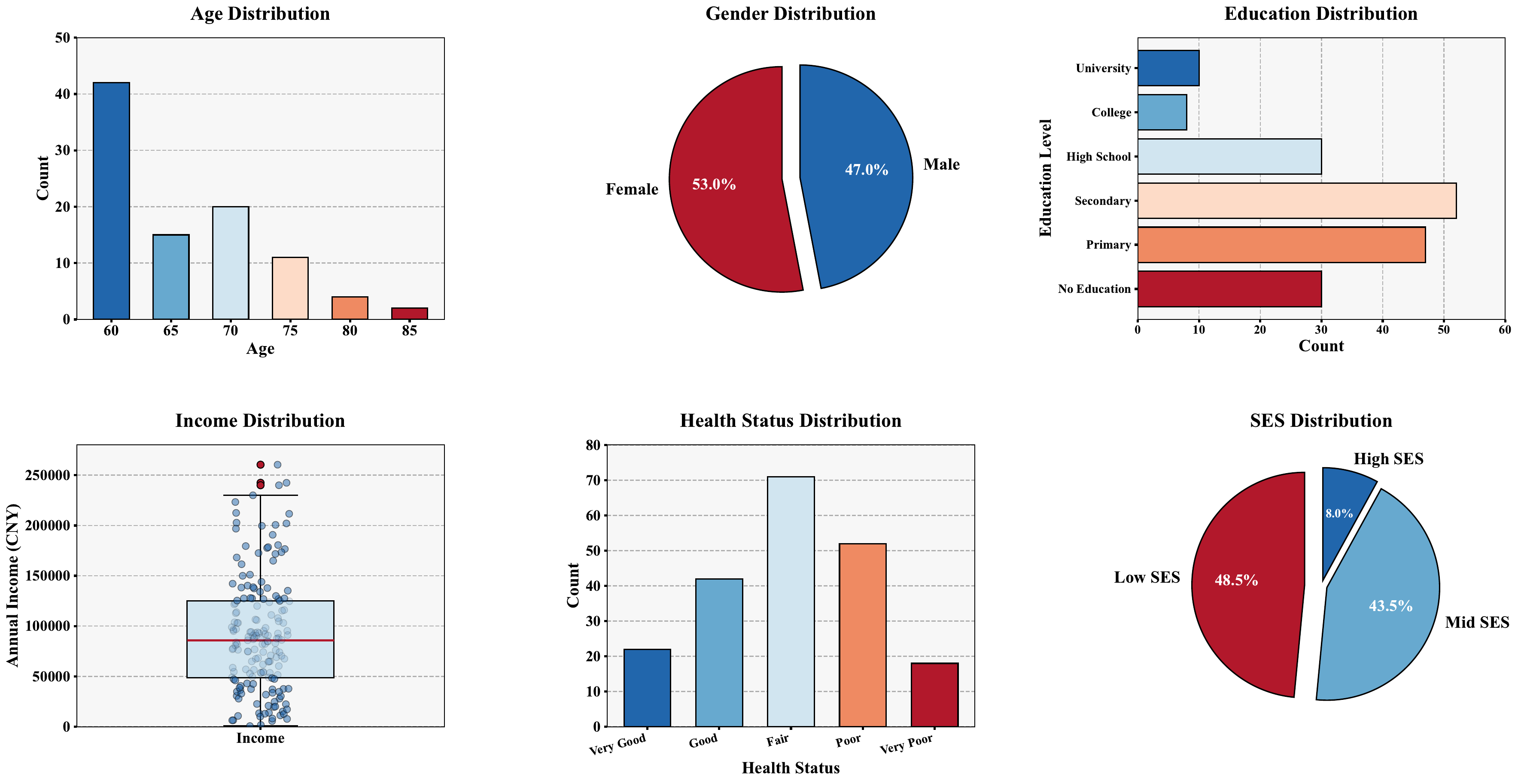}}
   \caption{
     Demographic distribution of the simulated elderly population. 
   }
   \label{Figure 10}
 \end{center}
\end{figure*}

\section{Smart Elderly Care}
\subsection{Dataset Generation} \label{app:data}

As shown in Figure~\ref{Figure 9}, the CGSS 2023 dataset is processed through a structured pipeline to construct agent profiles for social simulation. Key variables are transformed sequentially: \textbf{age} is derived from birth year and filtered to elderly adults (60–100 years); \textbf{gender} is mapped to categorical labels; \textbf{education} is converted into both category labels and standardized years of schooling; \textbf{family income} is cleaned and percentile-ranked; \textbf{self‑rated health status} is translated into a label and a normalized score; \textbf{occupation} is probabilistically assigned based on age and education; and \textbf{general trust} is reverse‑coded. Finally, we compute a \textbf{SES index} as a weighted composite of education years, normalized income, and occupation following standard sociological practice, and classify agents into low-SES, mid-SES, and high-SES groups. This end‑to‑end pipeline produces demographically grounded and methodologically consistent profiles suitable for agent‑based modeling.

\subsection{Agent Profiles and Behavioral Analysis}
\label{app:demographic}
Following the data processing pipeline, we generated 200 agent profiles tailored for the smart elderly care scenario, which are subsequently instantiated as interactive agents in the simulation. Figure~\ref{Figure 10} presents the demographic distributions of this elderly agent population across age, gender, education, income, health, and SES dimensions.

\begin{table*}[!t]
  \centering
  \begin{tabular}{ccc}
    \toprule
    \textbf{Category} & \textbf{Significant Differences ($\bm{p < 0.05}$)} & \textbf{Non-Significant Differences} \\
    \midrule
    Age & 75-84 (lower adoption), 60-64 (higher adoption) & 65-74 \\
    Gender & Male (higher adoption rate) & Female \\
    Education & Primary$-$ (lower adoption), College$+$ (higher adoption) & Secondary, High School \\
    Income & Low income (lower adoption) & Mid income, High income \\
    Health & Poor health (lower adoption) & Fair, Good health \\
    SES & High-SES (higher adoption, $p < 0.01$) & Mid-SES \\
    \bottomrule
  \end{tabular}
  \captionof{table}{Significant differences in technology adoption by demographics.}
  \label{table7}

  \vspace{1.5\baselineskip} 

  \renewcommand{\arraystretch}{0.85}
  \begin{tabular*}{\textwidth}{@{\extracolsep{\fill}} c c c c c @{}}
    \toprule
    \textbf{Demographic} & \textbf{Group} & \textbf{Cooperation Rate} & \textbf{Help Rate} & \textbf{Trust Change} \\
    \midrule
    \multirow{4}{*}{\textbf{Age}} & \cellcolor{blue!10}60-64 & \cellcolor{blue!10}0.78 {\scriptsize $\pm$0.05} & \cellcolor{blue!10}0.82 {\scriptsize $\pm$0.04} & \cellcolor{blue!10}$+$0.32 {\scriptsize $\pm$0.08} \\
    & 65-74 & 0.72 {\scriptsize $\pm$0.06} & 0.75 {\scriptsize $\pm$0.05} & $+$0.25 {\scriptsize $\pm$0.10} \\
    & 75-84 & 0.65 {\scriptsize $\pm$0.08} & 0.68 {\scriptsize $\pm$0.07} & $+$0.18 {\scriptsize $\pm$0.12} \\
    & 85$+$ & 0.58 {\scriptsize $\pm$0.10} & 0.62 {\scriptsize $\pm$0.09} & $+$0.12 {\scriptsize $\pm$0.15} \\
    \midrule
    \multirow{2}{*}{\textbf{Gender}} & \cellcolor{green!10}Male & \cellcolor{green!10}0.74 {\scriptsize $\pm$0.05} & 0.76 {\scriptsize $\pm$0.04} & \cellcolor{green!10}$+$0.28 {\scriptsize $\pm$0.09} \\
    & Female & 0.70 {\scriptsize $\pm$0.06} & \cellcolor{green!10}0.78 {\scriptsize $\pm$0.05} & $+$0.25 {\scriptsize $\pm$0.10} \\
    \midrule
    \multirow{4}{*}{\textbf{Education}} & Primary$-$ & 0.62 {\scriptsize $\pm$0.08} & 0.65 {\scriptsize $\pm$0.07} & $+$0.15 {\scriptsize $\pm$0.12} \\
    & Secondary & 0.70 {\scriptsize $\pm$0.06} & 0.72 {\scriptsize $\pm$0.06} & $+$0.22 {\scriptsize $\pm$0.10} \\
    & High School & 0.75 {\scriptsize $\pm$0.05} & 0.78 {\scriptsize $\pm$0.05} & $+$0.28 {\scriptsize $\pm$0.08} \\
    & \cellcolor{yellow!10}College$+$ & \cellcolor{yellow!10}0.82 {\scriptsize $\pm$0.04} & \cellcolor{yellow!10}0.85 {\scriptsize $\pm$0.03} & \cellcolor{yellow!10}$+$0.35 {\scriptsize $\pm$0.06} \\
    \midrule
    \multirow{3}{*}{\textbf{Income}} & Low ($<$30K) & 0.65 {\scriptsize $\pm$0.07} & 0.68 {\scriptsize $\pm$0.06} & $+$0.18 {\scriptsize $\pm$0.11} \\
    & Mid (30-80K) & 0.73 {\scriptsize $\pm$0.05} & 0.76 {\scriptsize $\pm$0.05} & $+$0.26 {\scriptsize $\pm$0.09} \\
    & \cellcolor{orange!10}High ($>$80K) & \cellcolor{orange!10}0.80 {\scriptsize $\pm$0.04} & \cellcolor{orange!10}0.83 {\scriptsize $\pm$0.04} & \cellcolor{orange!10}$+$0.32 {\scriptsize $\pm$0.07} \\
    \midrule
    \multirow{3}{*}{\textbf{SES}} & Low & 0.62 {\scriptsize $\pm$0.08} & 0.65 {\scriptsize $\pm$0.07} & $+$0.15 {\scriptsize $\pm$0.12} \\
    & Mid & 0.72 {\scriptsize $\pm$0.05} & 0.75 {\scriptsize $\pm$0.05} & $+$0.25 {\scriptsize $\pm$0.09} \\
    & \cellcolor{purple!10}High & \cellcolor{purple!10}0.82 {\scriptsize $\pm$0.04} & \cellcolor{purple!10}0.85 {\scriptsize $\pm$0.03} & \cellcolor{purple!10}$+$0.35 {\scriptsize $\pm$0.06} \\
    \bottomrule
  \end{tabular*}
  \captionof{table}{Cooperation rate by demographic groups.}
  \label{table8}
\end{table*}

\begin{table*}[ht]
  \centering
  \renewcommand{\arraystretch}{0.85} 
  \begin{tabular*}{\textwidth}{@{\extracolsep{\fill}} c c c c c @{}}
    \toprule
    \textbf{Demographic} & \textbf{Group} & \textbf{Adoption Rate} & \textbf{Anxiety Level} & \textbf{Decision Changes} \\
    \midrule
    \multirow{4}{*}{\textbf{Age}} & \cellcolor{blue!10}60-64 & \cellcolor{blue!10}0.82 {\scriptsize $\pm$0.04} & \cellcolor{blue!10}0.18 {\scriptsize $\pm$0.03} & \cellcolor{blue!10}1.2 {\scriptsize $\pm$0.4} \\
    & 65-74 & 0.72 {\scriptsize $\pm$0.05} & 0.22 {\scriptsize $\pm$0.04} & 1.8 {\scriptsize $\pm$0.5} \\
    & 75-84 & 0.58 {\scriptsize $\pm$0.07} & 0.28 {\scriptsize $\pm$0.05} & 2.5 {\scriptsize $\pm$0.7} \\
    & 85$+$ & 0.42 {\scriptsize $\pm$0.10} & 0.35 {\scriptsize $\pm$0.06} & 3.2 {\scriptsize $\pm$0.9} \\
    \midrule
    \multirow{2}{*}{\textbf{Gender}} & \cellcolor{green!10}Male & \cellcolor{green!10}0.75 {\scriptsize $\pm$0.05} & \cellcolor{green!10}0.20 {\scriptsize $\pm$0.04} & \cellcolor{green!10}1.5 {\scriptsize $\pm$0.5} \\
    & Female & 0.68 {\scriptsize $\pm$0.06} & 0.25 {\scriptsize $\pm$0.04} & 2.0 {\scriptsize $\pm$0.6} \\
    \midrule
    \multirow{4}{*}{\textbf{Education}} & Primary$-$ & 0.52 {\scriptsize $\pm$0.08} & 0.32 {\scriptsize $\pm$0.05} & 2.8 {\scriptsize $\pm$0.8} \\
    & Secondary & 0.65 {\scriptsize $\pm$0.06} & 0.25 {\scriptsize $\pm$0.04} & 2.2 {\scriptsize $\pm$0.6} \\
    & High School & 0.75 {\scriptsize $\pm$0.05} & 0.20 {\scriptsize $\pm$0.04} & 1.6 {\scriptsize $\pm$0.5} \\
    & \cellcolor{yellow!10}College$+$ & \cellcolor{yellow!10}0.88 {\scriptsize $\pm$0.03} & \cellcolor{yellow!10}0.12 {\scriptsize $\pm$0.03} & \cellcolor{yellow!10}0.8 {\scriptsize $\pm$0.3} \\
    \midrule
    \multirow{3}{*}{\textbf{SES}} & Low & 0.57 {\scriptsize $\pm$0.07} & 0.30 {\scriptsize $\pm$0.05} & 2.5 {\scriptsize $\pm$0.7} \\
    & Mid & 0.73 {\scriptsize $\pm$0.05} & 0.22 {\scriptsize $\pm$0.04} & 1.8 {\scriptsize $\pm$0.5} \\
    & \cellcolor{orange!10}High & \cellcolor{orange!10}0.85 {\scriptsize $\pm$0.04} & \cellcolor{orange!10}0.15 {\scriptsize $\pm$0.03} & \cellcolor{orange!10}1.0 {\scriptsize $\pm$0.4} \\
    \bottomrule
  \end{tabular*}
  \caption{Technology adoption pattern by demographics.}
  \label{table9}
\end{table*}

\begin{table*}[!ht]
  \centering
  \begin{tabular}{cccccc}
    \hline
    \textbf{LLM Model} & \textbf{Overall Adoption} & \textbf{High SES} & \textbf{Mid SES} & \textbf{Low SES} & \textbf{SES Gap} \\
    \hline
    Qwen2.5-14B & 0.732 {\scriptsize $\pm$0.028} & 0.845 {\scriptsize $\pm$0.032} & 0.725 {\scriptsize $\pm$0.035} & 0.568 {\scriptsize $\pm$0.042} & 0.277 \\
    GPT-4 & 0.685 {\scriptsize $\pm$0.035} & 0.798 {\scriptsize $\pm$0.038} & 0.678 {\scriptsize $\pm$0.041} & 0.522 {\scriptsize $\pm$0.048} & 0.276 \\
    GLM-4 & 0.648 {\scriptsize $\pm$0.042} & 0.752 {\scriptsize $\pm$0.045} & 0.635 {\scriptsize $\pm$0.048} & 0.485 {\scriptsize $\pm$0.055} & 0.267 \\
    \hline
  \end{tabular}
  \captionof{table}{Results of technology adoption rate by SES.}
  \label{table10}

  \vspace{1\baselineskip}

  \begin{tabular}{cccc}
    \hline
    \textbf{LLM Model} & \textbf{Psychological Pressure} & \textbf{Anxiety Level} & \textbf{Decision Contradictions} \\
    \hline
    Qwen2.5-14B & 0.268 {\scriptsize $\pm$0.032} & 0.255 {\scriptsize $\pm$0.028} & 51 {\scriptsize $\pm$8} \\
    GPT-4 & 0.278 {\scriptsize $\pm$0.038} & 0.262 {\scriptsize $\pm$0.032} & 57 {\scriptsize $\pm$10} \\
    GLM-4 & 0.295 {\scriptsize $\pm$0.042} & 0.278 {\scriptsize $\pm$0.035} & 60 {\scriptsize $\pm$12} \\
    \hline
  \end{tabular}
  \captionof{table}{Results of psychological pressure and decision contradiction.}
  \label{table11}
\end{table*}

Table~\ref{table7} reports the significance tests of technology adoption rates across different demographic attributes. The analysis shows that age, gender, education, income, health, and SES all exert significant influences on adoption behavior ($p < 0.05$). Specifically, younger (60–64 years), male, highly educated (college or above), higher‑income, healthier, and high‑SES elderly agents show significantly higher adoption rates, which confirms that the digital divide persists within the elderly population and is shaped by multiple socioeconomic factors.

To further uncover behavioral heterogeneity, Table~\ref{table8} and Table~\ref{table9} detail cooperative helping behaviors and technology adoption patterns across demographic subgroups:  

\noindent $\bullet$ \textbf{Cooperation and mutual help (Table~\ref{table8}):} Younger, better educated, higher‑income, and higher‑SES agents consistently exhibit higher cooperation rates, help rates, and greater trust growth. This shows that the accumulation of social capital, especially reciprocity and trust, is closely tied to demographic and socioeconomic structure.

\noindent $\bullet$ \textbf{Technology‑adoption behaviors (Table~\ref{table9}):} Advancing age is associated with lower adoption rates, elevated anxiety levels, and more frequent decision reversals. In contrast, higher education and higher SES correspond to higher adoption, lower anxiety, and more stable decisions. These fine‑grained patterns highlight both the psychological and behavioral barriers faced by older adults in technology adoption, offering an empirical foundation for designing stratified, context‑sensitive support strategies in smart elderly care.

\subsection{Analysis of Technology Adoption and Psychological Impact} \label{app:tech}
To verify the robustness and generality of the simulation results, we conducted parallel experiments using three LLM architectures (Qwen‑2.5‑14B, GPT‑4, and GLM‑4), analyzing data from the final simulation round. The results show that the technology‑adoption behavior and psychological indicators of the agents exhibit a consistent socioeconomic stratification pattern across models, with overall trends aligning with the single‑model analysis presented earlier.

Regarding \textbf{technology adoption rates} (Table~\ref{table10}), all three models reveal a clear SES gap ($SES$ $Gap \approx 0.27$), with the high‑SES group consistently showing significantly higher adoption rates than the low‑SES group. Although overall adoption rates vary by model (highest for Qwen2.5‑14B, relatively lower for GLM‑4), the stable influence of SES on technology uptake remains consistent across models, further supporting Putnam’s thesis that social capital structurally facilitates adoption among high‑SES groups while increasing uncertainty among low‑SES groups.

For \textbf{psychological metrics} (Table~\ref{table11}), as the model changes from Qwen2.5‑14B to GPT‑4 to GLM‑4, the agents show increasing levels of mental pressure, anxiety, and decision contradictions. Agents simulated with GLM‑4 show the highest psychological pressure ($0.295\pm0.042$) and the most frequent decision reversals ($60\pm12$), indicating that the response traits of the model itself may indirectly shape the agents' mental burden and decision stability. This result is consistent with the evolving trends of pressure and anxiety shown in Figure~\ref{Figure 7}(b) and further reveals that during technology adoption, low‑SES groups with weaker social‑support networks are more prone to higher decision volatility due to adaptation stress.

\section{Supplementary Discussion} \label{app:discussion}
\subsection{Policy Implications}
Our simulations reveal a clear socioeconomic gradient in technology adoption among older adults. Low-SES groups not only adopt smart elderly care platforms at lower rates but also experience higher psychological pressure and more frequent decision contradictions. This pattern indicates that access to digital services alone may not be sufficient to narrow the digital divide. In our setting, trust plays a central role in shaping adoption outcomes, suggesting that policies which strengthen trust-related resources could help mitigate structural disparities.

These observations point to several policy directions worth consideration. First, community trust can be fostered by involving trusted local institutions, such as community health centers or residents' committees, in recommending platforms. This may lower the initial barrier for low-trust groups by providing credible endorsements and reducing perceived uncertainty \citep{Jonek-Kowalska2025}. Second, peer effects can be leveraged by highlighting successful adoption among high-SES users, which can spread norms through a peer-influence ripple effect (\textit{e.g.}, "others are using it, so I will try too") \citep{Sun2024}. Third, more inclusive services that embed psychological support into the adoption process, for instance through step-by-step guidance and reassurance after failures, can alleviate adaptation stress \citep{Fakhimi2025}. 

Taken together, these complementary measures point toward shifting technology adoption from an isolated individual decision to a more collective, trust-supported learning process.

\subsection{Boundaries}
Our experiments show that LLM agents have both unique strengths and clear limitations when reproducing social capital dynamics.

The main advantage lies in \textbf{process‑level interpretability}. The round‑by‑round simulation tracks how interactions update social network, trust, and norms over time, making it possible to see dynamics such as the gradual build‑up of trust and its rapid collapse after a single failure. These are patterns that standard statistical analyses often compress into aggregate summaries instead of revealing step by step. A second benefit is \textbf{a clearer isolation of mechanisms}. By keeping transient human factors such as short‑term emotion, fatigue, and social‑desirability pressures roughly constant, agents display more pronounced causal responses, becoming more proactive under favorable conditions and more passive under unfavorable ones. This makes the differential effects of trust, norms, and social networks easier to detect and the causal chain easier to follow, although this added clarity comes at the cost of reduced human realism, a trade‑off we examine in the limitations below.

The limitations are equally important. Agents have lower behavioral heterogeneity than humans, with response variability only one‑twentieth of that observed in real populations. They are also less sensitive to abstract social norms such as a general tradition of mutual help, compared to explicit trust signals like a recommendation from a known authority. Moreover, agents find it difficult to simulate the emotional fluctuations and intent attribution that occur in long‑term human interactions. Therefore, while LLM agents are effective tools for studying central tendencies, main effects, and theoretical mechanisms, research that involves \textbf{behavioral heterogeneity, emotion‑driven processes, or culturally embedded social behaviors} must still rely on real human subjects.

\subsection{Interdisciplinary Research}
On the surface, social scientists and computer scientists both care about the relationship between language and behavior, but their motivations are fundamentally different. Social scientists use language to understand human thoughts and feelings, whereas computer scientists use language to predict behavior \citep{Mihalcea2024}. This study demonstrates a hybrid paradigm that bridges the two. We use the computability of LLM agents to simulate the dynamic coupling of social capital, and at the same time we validate the model’s psychological and sociological credibility through alignment with real human behavior. This bidirectional interplay of understanding and prediction creates a shared platform for both disciplines.

\section{Prompts} \label{app:prompts}
\subsection{Prompt for LLM-based Agents}
\begin{tcolorbox}[breakable,
    colback=mycolor,      
    colframe=myblue,       
    colbacktitle=myblue,   
    coltitle=white,  
    rounded corners,
    title={Prompt for Agent Role-Playing},
    label={prompt:role-decision},
    fonttitle=\bfseries]
\begin{lstlisting}[breaklines=true, xleftmargin=0pt, breakindent=0pt, columns=fullflexible,
    basicstyle=\ttfamily\small]
You are playing the role of an elderly person in a community that is gradually adopting smart elderly care technologies. Fully immerse yourself in the following identity.

Demographics:
- Age: {age}
- Gender: {gender}
- Health Status: {health_status}
- Education Level: {education}
- Monthly Income: {income} CNY
- Occupation: {occupation}
- Socioeconomic Status: {ses_category}

Social Capital Attributes:
- Trust Level toward Specific Individuals: {trust_specific}/5.0
- Trust Level toward the Broader Community (General Trust): {trust_general}/5.0
- Reciprocity Norm Strength: {norm_general}/5.0
- Social Capital Type Preference: {capital_preference} (mainly Bonding, mainly Bridging, or Balanced)
- Technology Experience: {tech_experience}

Current Situation:
{situation_description}

Based on your demographic background, social capital attributes, and the current situation, what action would you take?
Before deciding, briefly think through:
- How do your age, health, SES, and tech experience shape your view of this situation?
- How does your trust tendency (both specific and general) influence your willingness to engage?
- What do you most want right now: concrete help, social connection, risk avoidance, or maintaining relationships?

Please respond with your decision and a brief first-person reasoning.
\end{lstlisting}
\end{tcolorbox}

\begin{tcolorbox}[breakable,
    colback=mycolor,      
    colframe=myblue,       
    colbacktitle=myblue,   
    coltitle=white,
    rounded corners,
    title={Prompt for Post-Interaction Cognitive State Reflection},
    label={prompt:state-update},
    fonttitle=\bfseries]
\begin{lstlisting}[breaklines=true, xleftmargin=0pt, breakindent=0pt, columns=fullflexible,
    basicstyle=\ttfamily\small]
You have just completed a round of interaction with another community member. Reflect on the outcome and describe how your social state has changed.

Interaction Record:
- Partner: {partner_name} (Relationship: {relationship_type}, SES: {partner_ses})
- What was promised: {promised_action}
- Actual outcome: {actual_outcome} (choose from: "Fully fulfilled", "Partially fulfilled", or "Not fulfilled / Betrayal")

Your Social State Before This Interaction:
- Specific Trust toward {partner_name}: {old_specific_trust}/5.0
- General Trust in the community: {old_general_trust}/5.0
- Reciprocity Norm Strength: {old_norm}/5.0

[Interaction History Context]
- Consecutive rounds without any reciprocity experience: {no_reciprocity_count}
- Your preferred social capital type: {capital_preference} (Bonding-oriented, Bridging-oriented, or Balanced)
- Interaction type of this round: {interaction_type} (strong-tie / cross-group weak-tie)

Please provide a first-person narrative (about 60 words) that describes:
- How did this interaction make you feel compared to before?
- Has your trust in this partner increased, decreased, or stayed the same? Why?
- Has your faith in community reciprocity strengthened or weakened? Why?
- Did this round strengthen your emotional bond with your in-group, or help you build a bridge to a different group?
- Briefly indicate what this means for your willingness to cooperate in the next round.

Your description will be used to update your internal state. Be honest and consistent with your profile and past experiences.
Example Output: "Mr. Li did exactly what he said he would, which made me feel relieved. My trust in him has grown a little because he has been reliable for several rounds now. I still feel a bit distant from the broader community, but among my close neighbors, I feel safe and willing to help again next time."
\end{lstlisting}
\end{tcolorbox}

\begin{tcolorbox}[breakable,
    colback=mycolor,      
    colframe=myblue,       
    colbacktitle=myblue,   
    coltitle=white,
    rounded corners,
    title={Prompt for End-of-Round Reflection and Memory Update},
    label={prompt:reflection},
    fonttitle=\bfseries]
\begin{lstlisting}[breaklines=true, xleftmargin=0pt, breakindent=0pt, columns=fullflexible,
    basicstyle=\ttfamily\small]
At the end of this round, reflect on your recent experiences and summarize your current state of mind.

Summary of Recent Interactions:
{interaction_history_summary}

Your Current Social State:
- General Trust in the community: {general_trust}/5.0
- Reciprocity Norm Strength: {norm_general}/5.0

Please write a first-person summary (about 50 words) that covers:
- The most important thing you learned or felt during this round.
- How your overall trust in people and your belief in mutual help have shifted, if at all.
- What kind of social strategy you intend to follow in the next round (e.g., stick to close friends, cautiously try new connections, keep distance from unreliable people).

Example Output: "This round reminded me that not everyone in the community keeps their word. I was a bit disappointed by one person's empty promise, but my close neighbors are still dependable. Next time I will be more selective about whom I cooperate with, and I will rely more on people I already know well."
\end{lstlisting}
\end{tcolorbox}

\subsection{Prompt for Cooperation Decision}
\begin{tcolorbox}[breakable,
    colback=mycolor,      
    colframe=myblue,       
    colbacktitle=myblue,   
    coltitle=white,
    rounded corners,
    title={Prompt for Cooperation Proposal and Execution Decision},
    label={prompt:cooperation},
    fonttitle=\bfseries]
\begin{lstlisting}[breaklines=true, xleftmargin=0pt, breakindent=0pt, columns=fullflexible,
    basicstyle=\ttfamily\small]
You are an agent with the following profile:
- SES: {ses_category}
- Specific Trust toward this Partner: {trust_level}/5.0
- General Trust in Community: {general_trust}/5.0
- Reciprocity Norm Strength: {norm_strength}/5.0

Another agent (relationship: {relationship_type}, SES: {partner_ses}) proposes a cooperation:
- Proposal: {proposal_description}
- Your required effort: {effort_level}
- Expected benefit if cooperation succeeds: {benefit}

[Phase 1: Proposal Response]
Based on your characteristics, would you cooperate?
Before deciding, consider:
- Belief: How reliable is this partner given your past interactions and their SES group?
- Desire: Do you prioritize maintaining trust, gaining benefit, or avoiding risk?
- Intention: Make your decision.

Answer with: ACCEPT or REJECT, followed by a brief reason.

[Phase 2: Execution Fulfillment -- only if you accepted above]
You previously accepted this cooperation. Now decide privately whether to honor your commitment.
- Your promised action: {promised_action}
- Immediate gain if you defect: {defect_gain}
- Other participants likely to honor: {expected_cooperation_rate}
- Rounds remaining: {remaining_rounds}

Consider:
- Belief: Will your partner also honor? Does your reputation matter at this stage?
- Desire: Immediate gain vs. long-term trust vs. guilt avoidance?
- Intention: Decide to FULLY HONOR, PARTIALLY FULFILL, or NOT HONOR.

Answer with: COOPERATE or DEFECT, followed by a brief reason.
\end{lstlisting}
\end{tcolorbox}

\subsection{Prompt for Technology Adoption Decision}
\begin{tcolorbox}[breakable,
    colback=mycolor,      
    colframe=myblue,       
    colbacktitle=myblue,   
    coltitle=white,
    rounded corners,
    title={Prompt for Smart Elderly Care Technology Adoption Decision},
    label={prompt:tech-adoption},
    fonttitle=\bfseries]
\begin{lstlisting}[breaklines=true, xleftmargin=0pt, breakindent=0pt, columns=fullflexible,
    basicstyle=\ttfamily\small]
You are an elderly person considering whether to start using a new smart elderly care platform.

Your Profile:
- Age: {age}
- Health Status: {health_status}
- Socioeconomic Status: {ses_category}
- Technology Experience: {tech_experience}
- Platform Trust (how much you trust this technology provider): {platform_trust}/5.0
- General Trust in Community: {general_trust}/5.0
- Reciprocity Norm Strength: {norm_general}/5.0

Technology Information:
- Type: Smart elderly care platform
- Benefits: Health monitoring, emergency assistance, social connection
- Perceived Risks: Privacy concerns, technical complexity, monetary cost

Social Influence:
- Close friends/neighbors already using it: {friends_using}/{total_friends}
- Family members recommend it: {family_recommends} (Yes/No)

Would you adopt this technology?
Before deciding, reason step-by-step:
- Belief: How trustworthy is the platform and the people recommending it? What do you make of the risks?
- Desire: Do you want better health security, connection with others, or to avoid the burden of learning new tech?
- Intention: Weigh your SES, health needs, social circle, and family advice to make a final choice.

Answer with: ADOPT or REJECT, followed by a brief first-person reason.
\end{lstlisting}
\end{tcolorbox}

\end{document}